\documentclass[runningheads]{llncs}

\usepackage{graphicx}
\usepackage{amsmath,amssymb, bm}
\usepackage{color}
\usepackage{amssymb}
\usepackage{url}

\usepackage{bbold}
\usepackage[symbols,nogroupskip,record]{glossaries-extra}
\usepackage{subcaption}
\usepackage{cite}
\usepackage[ruled,vlined]{algorithm2e}
\usepackage[pagebackref,breaklinks,colorlinks]{hyperref}
\usepackage[capitalize]{cleveref}
\crefname{section}{Sec.}{Secs.}
\Crefname{section}{Section}{Sections}
\Crefname{table}{Table}{Tables}
\crefname{table}{Tab.}{Tabs.}

\newcommand{\diag}{\operatorname{diag}}

\newcommand{\argmin}{\operatorname*{arg\,min}}
\newcommand{\norm}[1]{\left\lVert#1\right\rVert}
\newcommand{\mat}[1]{\left[\begin{matrix}#1\end{matrix}\right]}

\newsavebox\mybox
\savebox\mybox{%
	\begin{minipage}[t]{0.48\linewidth}
		\includegraphics[width=0.75\linewidth]{example-image-a}
	\end{minipage}%
}
\newlength\ImageHt
\begin{document}

\title{GiNGR: Generalized Iterative Non-Rigid Point Cloud and Surface Registration Using Gaussian Process Regression}
\titlerunning{GiNGR}
\authorrunning{D. Madsen et al.}
\author{Dennis Madsen \and Jonathan Aellen \and Andreas Morel-Forster \and Thomas Vetter \and Marcel Lüthi\\{\tt\small \{dennis.madsen, jonathan.aellen, andreas.forster, thomas.vetter, marcel.luethi\}@unibas.ch}}
\institute{Department of Mathematics and Computer Science, University of Basel, \\Basel, Switzerland
}
\maketitle

\begin{abstract}
	In this paper, we unify popular non-rigid registration methods for point sets and surfaces under our general framework, GiNGR. GiNGR builds upon Gaussian Process Morphable Models (GPMM) and hence separates modeling the deformation prior from model adaptation for registration. In addition, it provides explainable hyperparameters, multi-resolution registration, trivial inclusion of expert annotation, and the ability to use and combine analytical and statistical deformation priors. But more importantly, the reformulation allows for a direct comparison of registration methods. Instead of using a general solver in the optimization step, we show how Gaussian process regression (GPR) iteratively can warp a reference onto a target, leading to smooth deformations following the prior for any dense, sparse, or partial estimated correspondences in a principled way. We show how the popular CPD and ICP algorithms can be directly explained with GiNGR. Furthermore, we show how existing algorithms in the GiNGR framework can perform probabilistic registration to obtain a distribution of different registrations instead of a single best registration. This can be used to analyze the uncertainty e.g. when registering partial observations. GiNGR is publicly available and fully modular to allow for domain-specific prior construction.
\end{abstract}

\section{Introduction}
 {\let\thefootnote\relax\footnote{{Code available at \href{https://github.com/unibas-gravis/GiNGR}{github.com/unibas-gravis/GiNGR}}}}
Registration of point clouds is used in many areas within computer vision, such as medical image analysis, simultaneous localization and mapping (SLAM) to analyze an agent's environment within robotics and Geoscience. Closely related is also surface or volume registration where the points have additional connectivity information such as lines, triangles, or tetrahedrons. 
Within the last 20+ years, a vast majority of different registration algorithms have been published, which all try to find an optimal transformation between a reference (sometimes referred to as template or source) to a target. Many of the most popular point registration methods are summarized in \cite{maiseli_recent_2017, zhu_review_2019}. However, they do not include surface registration methods such as \cite{feldmar_rigid_1996, granger_multi-scale_2002, allen_space_2003, amberg_optimal_2007, zou_non-rigid_2007, cheng_statistical_2017, liang_nonrigid_2018, madsen_closest_2020}. With so many different methods, it can be difficult to know the difference between all of them and therefore also which one to choose for a specific problem. 

In this paper, we focus on non-rigid registration between a single reference and a target. This is also known as pairwise registration \cite{zhu_review_2019}. For simplicity, we therefore always refer to the non-rigid version of an algorithm, unless explicitly being stated otherwise. We do not consider many of the recent state-of-the-art methods based on neural networks as they often need a large dataset to learn the deformation space from and can therefore be categorized as groupwise registration. An example is \cite{jiang_disentangled_2020} which is the current best-ranked method for the FAUST dataset challenge \cite{bogo_faust_2014}. Our framework requires no training data to learn its deformations from. However,  within the same framework, we can utilize statistical deformation priors learned from data and even combine them with analytically defined deformation priors.

Three main assumptions define non-rigid registration:
\begin{enumerate}
	\item Regularization - How similar should deformation of neighboring points be?
	\item Correspondence - How to estimate corresponding point pairs between the reference and the target?
	\item Robustness - What is the noise assumption of the observed correspondence pairs? And how are "bad" correspondence pairs handled?
\end{enumerate}
Choosing the three assumptions differently leads to a large number of different methods. But by clearly defining them, and unifying registration methods in the same framework, we are able to obtain intuitive hyperparameters and fairly compare algorithms. 

Among other explanations, we think the number of registration methods is that high because they are often tuned to specific datasets and because their code is often not available. The hyperparameters are also often difficult to interpret and thereby difficult to choose optimally. Within our open-source framework, we have implemented a modular setup to easily switch between different registration strategies. Hence we can make algorithms easier comparable and adaptable to new tasks.

The basics of the Generalized iterative Non-Rigid Registration using Gaussian Process Regression (GiNGR) framework have informally been around for years. A minimal non-rigid ICP version was introduced together with Gaussian Process Morphable Models (GPMMs) \cite{luthi_gaussian_2017} in a MOOC \footnote{\href{https://www.futurelearn.com/courses/statistical-shape-modelling}{futurelearn.com/courses/statistical-shape-modelling}}. In this paper, we generalize the steps from the MOOC and show how this generalization makes it possible to reformulate many existing algorithms into the modular GiNGR framework. Our implementation of the registration framework works both for point sets and different mesh types such as triangulated and tetrahedral meshes.

Our contributions in this paper are:
\begin{itemize}
	\item We introduce the GiNGR framework for non-rigid point-set/surface registration in \cref{sec:method}.
	\item We reformulate the Iterative Closest Point (ICP) and Coherent Point Drift (CPD) algorithm and show that they are special instances under GiNGR in \cref{sub:icp_cpd}.
	\item We show how GiNGR allows for probabilistic registration in \cref{subsub:deterministic-vs-probabilistic}.
\end{itemize}
\section{Background}\label{sec:background}
In this section, we formalize the non-rigid registration task and we introduce GPR and GPMMs.

The goal of non-rigid registration is to deform a reference surface $\Gamma_R$ sampled at $n$ points $X_R \in \mathbb{R}^{n \times d}$ in $d$ dimensions to a target surface $\Gamma_T$ sampled at $m$ points $X_T \in \mathbb{R}^{m \times d}$. What we are looking for is a deformation vector for each point in $X_R$, stored in the matrix $U \in \mathbb{R}^{n \times d}$, such that $\tilde{U} = \argmin_{U} \mathbb{d}\left[\Gamma_R+U, \Gamma_T\right] + \lambda R(U)$. We denote regularized deformations as $\tilde{U}$ while we write $\hat{U}$ for the observed deformations.

Gaussian Process Morphable Models (GPMMs) \cite{luthi_gaussian_2017} use Gaussian Processes (GPs) to describe deformations from a reference point-set. This formalization is a generalization of the classical Point Distribution Model (PDM) \cite{cootes_training_1992}. A Gaussian process models deformations as a continuous function $u \sim \mathcal{GP}(\mu,k)$, completely defined by its mean function $\mu: \Gamma_R \rightarrow \mathbb{R}^d$ and a kernel function $k: \Gamma_R \times \Gamma_R \rightarrow \mathbb{R}^{d \times d}$. An extensive overview of the area of GPs can be found in \cite{rasmussen_gaussian_2006}. 
From a reference shape $\Gamma_R \subseteq \Omega$ with $\Omega \subset \mathbb{R}^d$ being the domain, the deformations that the reference shape can undergo is distributed according to the defined GP $u\sim \mathcal{GP}(\mu, k)$, with a shape defined such that
\begin{equation}
	\Gamma=\{ x + u(x) | x \in \Gamma_R\}
\end{equation}

As we are usually interested in smooth deformations, a sufficiently accurate low-rank approximation of the GP can be found where $r \ll n$:
\begin{equation}
	u[\bm{\alpha}](x)=\mu(x)+\sum_{i=1}^{r}\bm{\alpha}_i \sqrt{\lambda_i}\phi_i(x), \bm{\alpha}_i \sim \mathcal{N}(0,1)
\end{equation}
Using the low-rank representation (with $\lambda_i$ being the variance associated with the $i$'th basis function $\phi_i$), any deformation $u\in U$ is uniquely determined by the coefficient vector $\bm{\alpha} \in \mathbb{R}^r$:
\begin{equation}
	\Gamma[\bm{\alpha}] = \{ x + \mu(x) + \sum_{i=1}^{r} \bm{\alpha}_i \sqrt{\lambda_i}\phi_i(x) | x \in \Gamma_R \}.
\end{equation}
By formulating GiNGR using GPMMs, we are able to reduce the computational costs of registering large point sets by using $r\ll n$.

In \cref{sub:icp_cpd} we derive how the kernel function $k$ has been defined in popular algorithms such as CPD \cite{myronenko_point_2010} and ICP algorithms \cite{amberg_optimal_2007}. Besides being able to analytically define the kernel function, the deformation prior can also be learned from a set of example shapes in correspondence \cite{luthi_gaussian_2017}. 

As we are only interested in deformations at a discrete set of locations, we can discretize the continuous GP. For the observed deformations $\hat{U}$, and the predicted deformations $\tilde{U}$ the joint distribution according to the prior can then be written as
\begin{equation} 
    \left[
    \begin{matrix}
        \hat{U}\\
        \tilde{U}
    \end{matrix}    
    \right] \sim \mathcal{N}\left( 
    \left[
    \begin{matrix}
        \Mu_{X}\\
        \Mu_{X*}
    \end{matrix}  
    \right]
    , 
    \left[
    \begin{matrix}
        K(X,X)+\sigma^2I_{n} & K(X,X_*))\\
        K(X_*,X) & K(X_*,X_*)\\
    \end{matrix}
    \right]
    \right)
\end{equation}
Given $n$ observed and $n_*$ predicted deformations, $K(X,X)$ is the $n \times n$ covariance matrix of the observed deformations, $K(X,X_*)$ is the $n \times n_*$ matrix of covariances evaluated at all pairs of observed and predicted deformations and similarly for $K(X_*,X_*)$ and $K(X,X_*)$. For every observed deformation, we assume independent and identically distributed Gaussian noise. The mean of the posterior distribution can be computed in closed form as:
\begin{equation}\label{eq:gp-reg-mean-full}
	\tilde{\Mu} = \Mu_{X*} + K(X_*,X)(K(X,X)+\sigma^2 I_n)^{-1}\hat{U}.
\end{equation}
Note, that this leads to a full deformation field, even when the deformations are observed only partially, e.g. when only partial correspondences are predicted for a partially observed target. We refer to the closed-form continuous posterior model obtained with GPR as $\mathcal{GP}_p$. 

If the observed and predicted deformations are the same, then the mean of GPR simplifies to
\begin{equation}\label{eq:gp-reg-mean-simple}
	\tilde{\Mu} = \Mu_{X} + K(K+\sigma^2 I_n)^{-1}\hat{U},
\end{equation}
with $K$ being a shorthand notation for the covariance matrix spanned by the observed deformations $K(X,X)$.

\section{Method}\label{sec:method}
GiNGR is a generalization of existing non-rigid registration methods. We follow the iterative optimization method from \cite{amberg_optimal_2007}, where the correspondence is recomputed and fixed for every iteration. A property of iteratively re-estimating the correspondence points is that the optimization can be non-monotonically decreasing, which can help avoid getting stuck in local optima, as highlighted in \cite{amberg_optimal_2007}. Besides being able to generalize existing deterministic registration algorithms, GiNGR also facilitates probabilistic registration as introduced in \cite{morel-forster_probabilistic_2018, madsen_closest_2020}. In probabilistic registration, a stochastic step is taken in each iteration. Instead of finding the "best" registration, the objective is instead to approximate the posterior distribution
\begin{equation} 
P(\bm{\alpha}|\Gamma_T)\sim P(\Gamma_T|\bm{\alpha})P(\bm{\alpha}).
\end{equation}
With a GPMM as the prior model, $\bm{\alpha}$ follows a standard multivariate normal distribution and is therefore easily evaluated. For the likelihood function, the independent point likelihood is introduced in \cite{morel-forster_probabilistic_2018} which evaluates the distances between closest points under a normal distribution with a user-defined variance as a hyperparameter. GiNGR uses the principles from the informed ICP proposal introduced in \cite{madsen_closest_2020} and generalizes it to be able to use different correspondence and uncertainty estimation functions.

\begin{figure}[t]
	\centering
		\includegraphics[trim={50px 120px 50px 70px},clip, width=1.0\linewidth]{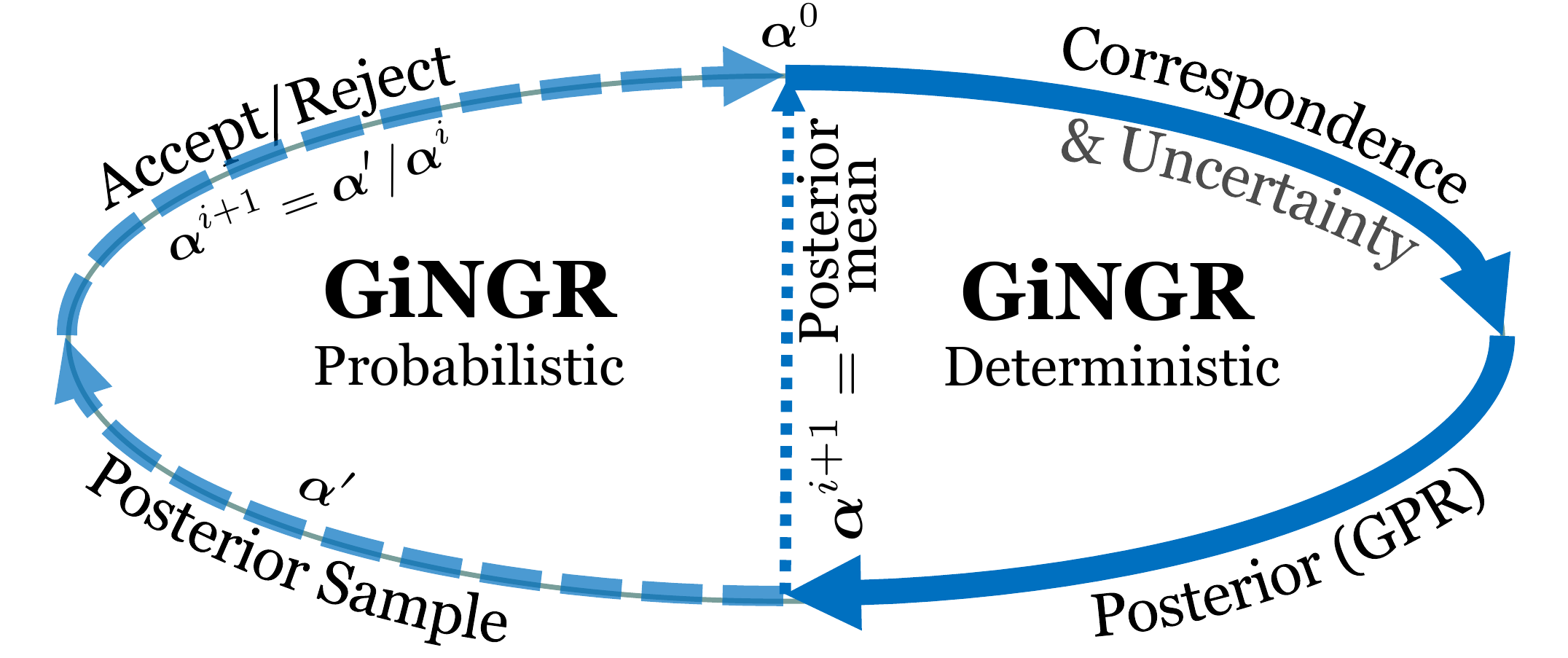}
	\caption{Visualization of a GiNGR iteration. Before the registration begins, the GPMM model parameters are initialized. Then the correspondence pairs between the reference and the target are estimated together with their associated uncertainty. Then $\mathcal{GP}_p$ is computed using GPR. Deterministic mode: the mean from $\mathcal{GP}_p$ is chosen. Probabilistic: a random sample from $\mathcal{GP}_p$ is proposed and either accepted or rejected based on the computed acceptance probability. Deterministic mode: ends when convergence is reached. Probabilistic mode: continues exploring the posterior until max-iteration.}
	\label{fig:gingrVisual}
\end{figure}

In \cref{fig:gingrVisual} we show the main steps of GiNGR. For the probabilistic setting, the posterior distribution $P(\bm{\alpha}|\Gamma_T)$ is estimated using the Metropolis-Hastings (MH) algorithm \cite{robert_monte_2013}. 
As a minimal setup, one can use the Gaussian kernel
\begin{equation}\label{eq:gaussianKernel}
	k(x,x')=\exp \left(\frac{-||x-x'||^2}{2\beta^2} \right),
\end{equation}
in the GPMM, with $\beta$ set to the wanted point correlation distance. The uncertainty $\sigma^2$ from \cref{eq:gp-reg-mean-full} and \cref{eq:gp-reg-mean-simple} is the estimated correspondence uncertainty, which can be manually set based on the noise assumption for the estimated correspondences.
The correspondence deformations in $\hat{U}$ can be computed using closest point estimate. The $\bm{\alpha}$ parameters are then updated based on the $\mathcal{GP}_p$ model according to \cref{fig:gingrVisual}. 

The uncertainty $\sigma^2$ is either manually or automatically decreased in each iteration as the reference gets closer to the target. An automatic update strategy is seen in \cite{myronenko_non-rigid_2007}, where gradual annealing is used to decrease $\sigma^2$ in each step, or \cite{myronenko_point_2010, hirose_bayesian_2020}, where the updated uncertainty is computed based on the variance difference between the reference and target points and the deformation update.

Seen from a practical aspect when registering two surfaces using the GiNGR implementation, one starts out with defining the deformation prior for the specific application. The GPMM is then constructed for a compact description of the model. We can sample and visualize the deformation prior separately from the registration \cite{luthi_gaussian_2017}. For the registration, the correspondence estimating function needs to be chosen (e.g. probabilistic or heuristic closest points) and a strategy to update the uncertainty in each iteration.

\subsection{Unifying ICP and CPD}\label{sub:icp_cpd}
In this section, we show how existing methods are special instances in GiNGR. The notation has been modified slightly to allow for an easier connection between the different algorithms. We also provide links to the corresponding equations in the original papers for reference.

\subsubsection{Coherent Point Drift}\label{sub:cpd}
CPD was originally introduced as a non-rigid point-set registration \cite{myronenko_non-rigid_2007}. Later, CPD was generalized to allow for rigid and affine transformations as well as automatically compute some of the hyperparameters \cite{myronenko_point_2010}.
What we are searching for in CPD is the deformation vectors $\tilde{U}$ from the reference point-set $X_R$ such that a new point-set location is found in each iteration $\hat{X}_R=X_R+\tilde{U}$. Initially $\hat{X}_R=X_R$. The deformation is iteratively computed based on the new point-set location $\hat{X}$ until the distance between the reference and the target surface meets a predefined convergence threshold. The CPD algorithm uses an expectation-maximization (EM) approach.

In the E step, a probabilistic correspondence matrix $P$ is computed \cite[Eq. 6]{myronenko_point_2010}, where each entry is defined as:
\begin{equation}
    p_{ij}=\frac{k(\hat{X}_R^i, X_T^j)}{\sum_{ii=1}^{m}k(\hat{X}_R^{ii}, X_T^j)+C(w)}
\end{equation}
with $k$ being a Gaussian kernel, $\hat{X}_R^i$ $X_T^j$ being the $i'th$ and $j'th$ point on $\hat{X}_R$ and $X_T$ respectively and $C(w)$ being an outlier distribution controlled by $0\leq w \leq 1$ which can be seen as the percentage of outlier points that the user can set manually. Let us consider $w=0$, then each column of the $P$ matrix sums to $1$, each entry in the matrix in other words states the probability of the $j$'th point in the target point-set corresponding to the $i$'th point in the reference point-set. 
In the following, we will also be using the $P\bm{1}$ vector, which is the row-wise sum of the $P$ matrix. These values give some direct insight into the confidence of the correspondences. A higher value will give more confidence to the deformation of this point from the reference and translate into a lower uncertainty for the correspondence pair in GiNGR.

In the M step, the probability matrix is kept fixed and the MAP deformation is computed. In CPD the deformation vectors are defined with $\tilde{U}=K_G W$. $K_G$ is a Gaussian kernel matrix with entries $k_{ij}$ according to \cref{eq:gaussianKernel}. $W$ is the weighted sum of deformations between a point on the reference and all the points on the target
$$w_n=\frac{1}{\sigma^2*\lambda}\sum_{n=1}^{}P(x_n|y_m)(y_m-\hat{x}_n).$$
The computation of $W$ can be rewritten \cite[Eq. 2]{myronenko_point_2010} to
\begin{equation}
	K_GW=K_G(K_G+\lambda \sigma^2 Q)^{-1}(QPX_T-\hat{X}_R)
\end{equation}
with $Q=\diag(P\bm{1})^{-1}$ being the diagonal matrix formed using the vector $P\bm{1}$. By comparing this formulation to \cref{eq:gp-reg-mean-simple}, we have that $K=K_G$, $\sigma^2 I_n=\lambda\sigma^2 Q$ and 
\begin{equation}\label{eq:cpd-observation}
	\hat{U}=QPX_T-\hat{X}_R
\end{equation}
One iteration of CPD is in other words equivalent to one iteration of GPR where the width of the kernel of the $\mathcal{GP}$ is specified from the $\beta$ hyperparameter. Each deformation observation in $\hat{U}$ is a simple sum of all possible correspondence pairs where each pair is scaled with their correspondence probability. We can also see that the noisy observation is scaled with the inverse of the $P\bm{1}$ entries. So a high $P\bm{1}$ value leads to a lower noise assumption on that specific observation. The $\lambda$ hyperparameter is used as a simple manual scaling of the noise observation. 

\subsubsection{Bayesian Coherent Point Drift}
In \cite{hirose_bayesian_2020}, the original CPD was generalized using Variational Bayesian Inference (VBI) formulation, calling it Bayesian Coherent Point Drift (BCPD). The BCPD algorithm follows the original CPD paper in large parts, we therefore only mention the main differences to the original paper. The reformulation to BCPD includes an optimization scheme for non-Gaussian kernels, to which the original CPD algorithm is limited to. The reformulation also includes integration of combined optimization of the global similarity transform and the local non-rigid deformations. The work was then extended and called BCPD++, in order to speed up the BCPD algorithm using GPR to allow for down-sampling of the point sets \cite{hirose_acceleration_2020}. As we have shown in \cref{sub:cpd}, CPD can be reformulated as GPR which is also the case for BCPD. In GiNGR, the two-stage BCPD++ algorithm can instead be unified under a single $\mathcal{GP}$ instead of having one $\mathcal{GP}$ for the BCPD step and one for the interpolation steps. 

In the BCPD formulation, the point sets are vectorized, so $X_R \in \mathbb{R}^{3n}$. The updated point-set in each iteration is:
\begin{equation}
    \hat{\bm{x}}_R = \mathcal{T}(\bm{x}_R + \tilde{\bm{u}}) = s(I_m \otimes R)(\bm{x}_R + \tilde{\bm{u}}) + (1_m \otimes t)
\end{equation}
with $\mathcal{T}$ being the similarity transformation with a scaling $\mathbb{s} \in \mathbb{R}$, rotation $R\in \mathbb{R}^{d\times d}$ and translation $t\in \mathbb{R}^{d}$ and $\otimes$ being the Kronecker product.
The correspondence matrix $P$ is largely the same as in the standard CPD, but with a few improvements. The matrix is updated with an indicator function to note if a point is an outlier as well as having updated the outlier distribution for which the integral approaches $0$ when $n$ becomes large in the standard CPD \cite[Eq. 8]{hirose_bayesian_2020}.

In comparison to \cref{eq:cpd-observation}, the local deformation observations in BCPD are:
\begin{equation}\label{eq:bcpd-observation}
    \hat{U}= (\mathcal{T}^{-1}(Q P X_T) - X_R)
\end{equation}
which can be explained as the local deformations modeling the residual of the global similarity transformation $T$. 
The regularized deformations are then
\begin{equation}
    \tilde{U}=K_GW=K_G(K_G+\lambda \frac{s^2}{\sigma^2} Q)^{-1} (\mathcal{T}^{-1}(Q P X_T) - X_R)
\end{equation}
Finally, the updated global similarity transformation is a least-squares estimation of transformation parameters between two point sets, similar to \cite{umeyama_least-squares_1991}. The point sets used to compute the transformation are the scaled target-points $Q P X_T$ and the reference points updated with the current local deformation $X_R+\tilde{U}$. 

\subsubsection{Iterative Closest Point - Translations}
ICP contains a whole range of different methods, where the task is to find the transformation that best aligns a reference to a target. The transformation is found iteratively, based on a set of correspondences found from closest point searching in each step. The transformations can be rigid \cite{besl_method_1992, chen_object_1992} or non-rigid deformation \cite{feldmar_rigid_1996, allen_space_2003, amberg_optimal_2007, cheng_statistical_2017, liang_nonrigid_2018}. Here, we look at ICP-T \cite{amberg_optimal_2007}, which is a non-rigid registration method for 3-dimensional surfaces. The ICP-T algorithm is a simple reformulation of \cite{allen_space_2003}, which iteratively finds the local translation deformation for each vertex in the point-set. In ICP-T, a stiffness term is introduced to regularize how much neighboring vertices are allowed to differ in their deformations. 

ICP-T minimizes a standard quadratic cost function:
\begin{equation}
	\argmin\limits_\mathrm{x} = \norm{ \mathcal{A}\mathrm{x}- \mathcal{B} }_F^2,
	\label{eq:quadraticproblem}
\end{equation}
with the terms
\begin{equation}
	\argmin_{\tilde{U}} = \norm{
		\mat{ \lambda_s B \\ WI} \tilde{U} - \mat{ 0 \\ W(X_c-X_R) }
	}_F^2
	\label{eq:deformationsquadratic}
\end{equation}
where $\lambda_s$ is a stiffness parameter governing the regularization strength of the deformations, $W\in \mathbb{R}^{n \times n}$ being a diagonal weight matrix identifying robust correspondence points, $B \in \mathbb{R}^{r \times n}$ being the incidence matrix and $r$ the number of edges in the reference mesh. The closest point location for each of the points in $X_R$ are contained in $X_c \in \mathbb{R}^{n \times d}$. For simplicity in the derivation, we set $W=I$ and $\lambda_s=1$. Solving \cref{eq:quadraticproblem} using least-squares leads to
\begin{equation}
	\mathrm{x}=(\mathcal{A}^{T}\mathcal{A})^{-1}\mathcal{A}^T\mathcal{B}
\end{equation}
Using the values from \cref{eq:deformationsquadratic}
\begin{equation}
	\tilde{U}=(B^{T}B+\lambda_s^2 I)^{-1}(X_c-X_R)    
\end{equation}
Here we recognize that $L=B^{T}B$ is the Laplacian matrix. We now make use of the Woodbury matrix identity for invertible matrices \cite{woodbury_inverting_1950}:
\begin{equation}
	(\mathcal{A}+\mathcal{U}\mathcal{C}\mathcal{V})^{-1} =
	\mathcal{A}^{-1}-\mathcal{A}^{-1}\mathcal{U}(\mathcal{C}^{-1}+\mathcal{V}\mathcal{A}^{-1}\mathcal{U})^{-1}\mathcal{V}\mathcal{A}^{-1}
\end{equation}
with $\mathcal{A}^{-1}\sim L^{\dagger}=K$, all $\mathcal{U}$,$\mathcal{C}$, and $\mathcal{V}$ chosen as identity matrices as well as $\hat{U}=X_c-X_R$, we can show that ICP-T chooses the deformations as
\begin{equation}
	\tilde{U}=Z+K(K+\lambda_s^2 I)^{-1}\hat{U}.
\end{equation}
The Laplacian matrix is not invertible as one of its eigenvalues is $0$. Instead, the generalized inverse (pseudo-inverse) is used which is why a correction term $Z$ is added. The general inverse of the Laplacian matrix and its applications are summarized in \cite{gutman_generalized_2004}. The correction term of the generalized inverse Laplacian is discussed in more detail in \cref{subsub:invlap-correction}.

More details about the intermediate steps of the derivation can be found in \cref{app:icp-t}.

To identify outlier points and remove them from the cost function, a weighting matrix is included in the optimization \cite{amberg_optimal_2007}. In GPR, non-robust points are directly removed from the observation data $\hat{U}$. In \cref{appsub:weight-matrix} we include the derivation showing that this is equivalent to what is done in the ICP methods. In \cref{appsub:weight-matrix} we also show that $\lambda_s$ from \cref{eq:deformationsquadratic} is equivalent to $\sigma$ from \cref{eq:gp-reg-mean-simple} which is the independent Gaussian noise assumption per observation. 

\subsubsection{Iterative Closest Point - Affine transformations}
We now turn to the ICP-A algorithm \cite{amberg_optimal_2007}, which models local differences as affine transformations, and should be more robust in registering partial data. The final transformation matrix $\mathcal{M} \in \mathbb{R}^{4n \times 3}$ no longer consist of simple deformations, but instead of an affine transformation matrix for each point in the reference. 
\begin{equation}
	\argmin_{\mathcal{M}} = \norm{
		\mat{ \lambda_s B \otimes G \\ WD} \mathcal{M} - \mat{ 0 \\ WX_c }
	}_F^2
\end{equation}
with $G$ being the $\diag([ 1,1,1,\gamma ]^T)$ and $\gamma$ depending on the units of the data and $D\in \mathbb{R}^{4n\times 3}$ being the sparse mapping of $X_R$. Again, we begin with the solution to the least-squares problem. For simplicity, we have set $W=I$ and $\lambda_s=1$ as in the ICP-T example. 
\begin{equation}
	\mathcal{M}=( B^{T}B \otimes G + D^TD)^{-1}(X_c) 
\end{equation}
The full derivation can be found in \cref{app:icp-a}, which follows the same procedure as ICP-T. For ICP-A we end up with the kernel being $\mathcal{K}=DKD^T$ with $K$ being the generalized inverse Laplacian matrix and $DD^T$ being the dot product kernel \cite{rasmussen_gaussian_2006}, which is what makes the ICP-A algorithm invariant to rotations around the origin. 
To keep the observations as simple deformations, they are corrected to adjust for the modeling of affine transformations $\hat{U}=X_c - (X_R+ \mathcal{K}^{-1}X_R)$.

Many papers exist which expand on \cite{amberg_optimal_2007} for their specific domains. One paper is \cite{hontani_robust_2012}, where they learn a Statistical Shape Model (SSM) from a training set over the possible affine transformations. In \cref{sub:pdm}, we describe how GiNGR similarly allows for the usage of a statistically learned kernel.
\begin{figure}[t]
	    \rotatebox[origin=c]{90}{\makebox[\ImageHt]{Pose 1}}\hspace{0.1em}
		\begin{subfigure}{0.23\linewidth}
        	\centering
			\includegraphics[width=\linewidth]{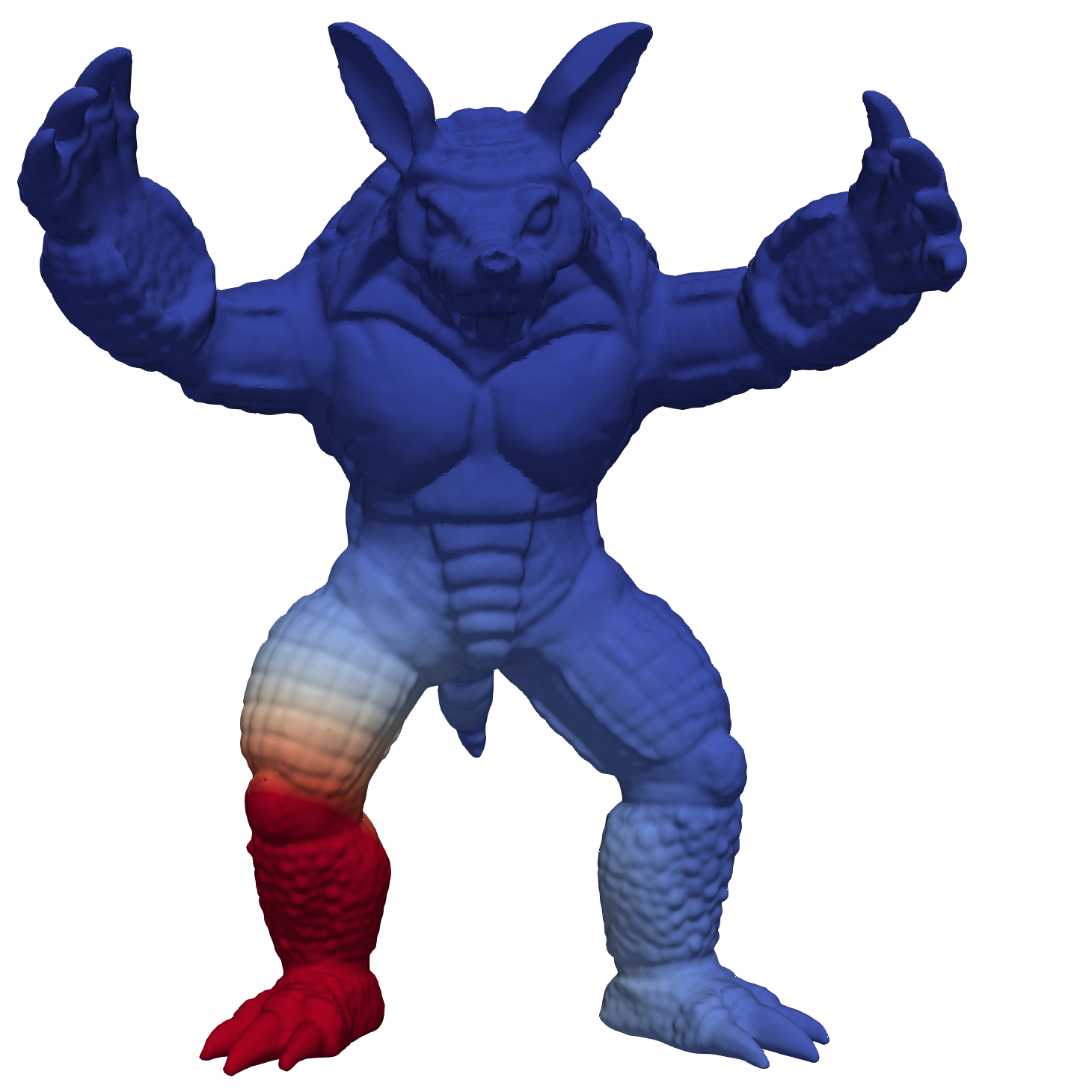}
		\end{subfigure}
		\begin{subfigure}{0.23\linewidth}
			\centering
			\includegraphics[width=\linewidth]{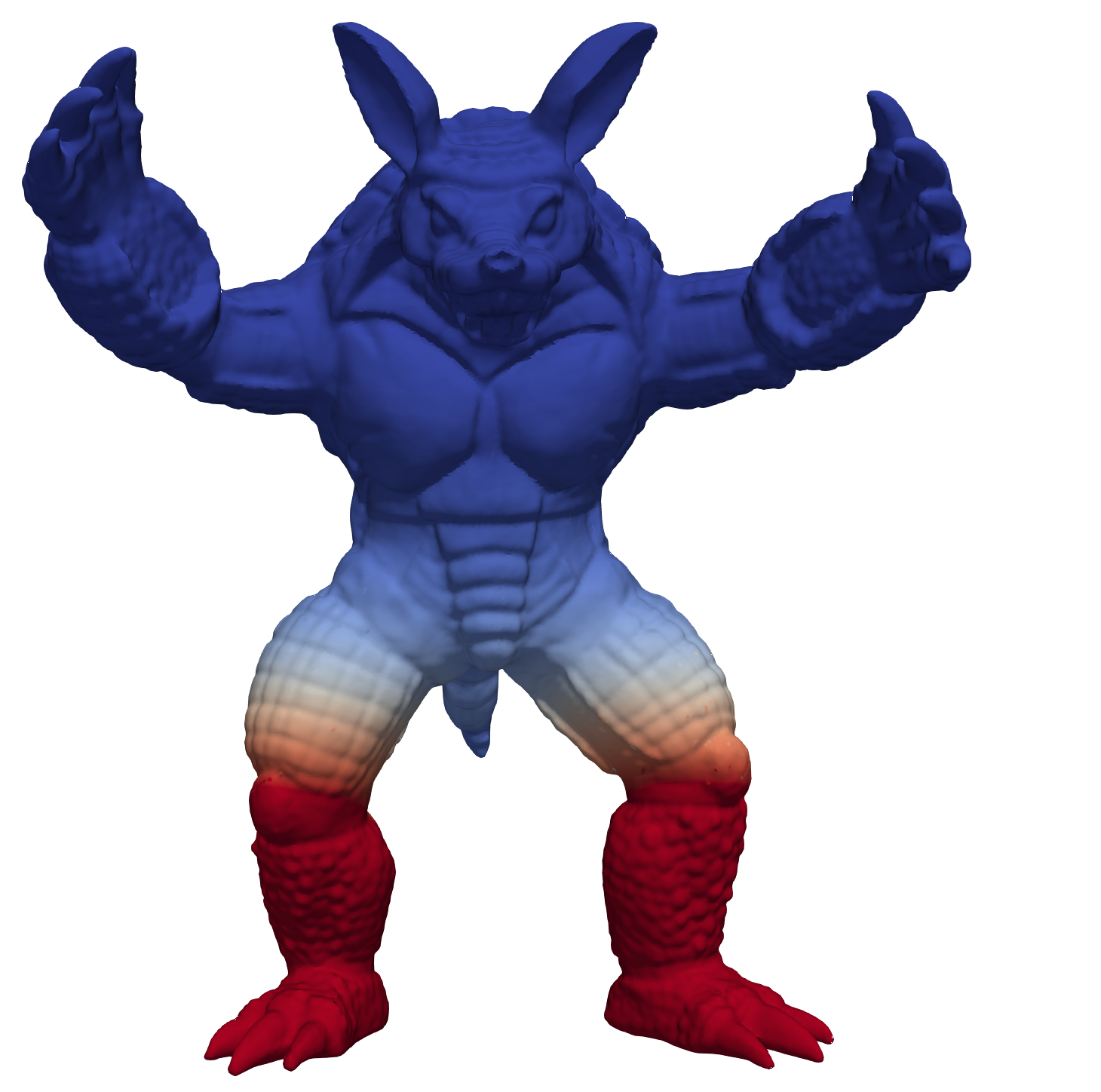}
		\end{subfigure}
		\begin{subfigure}{0.23\linewidth}
			\centering
			\includegraphics[width=\linewidth]{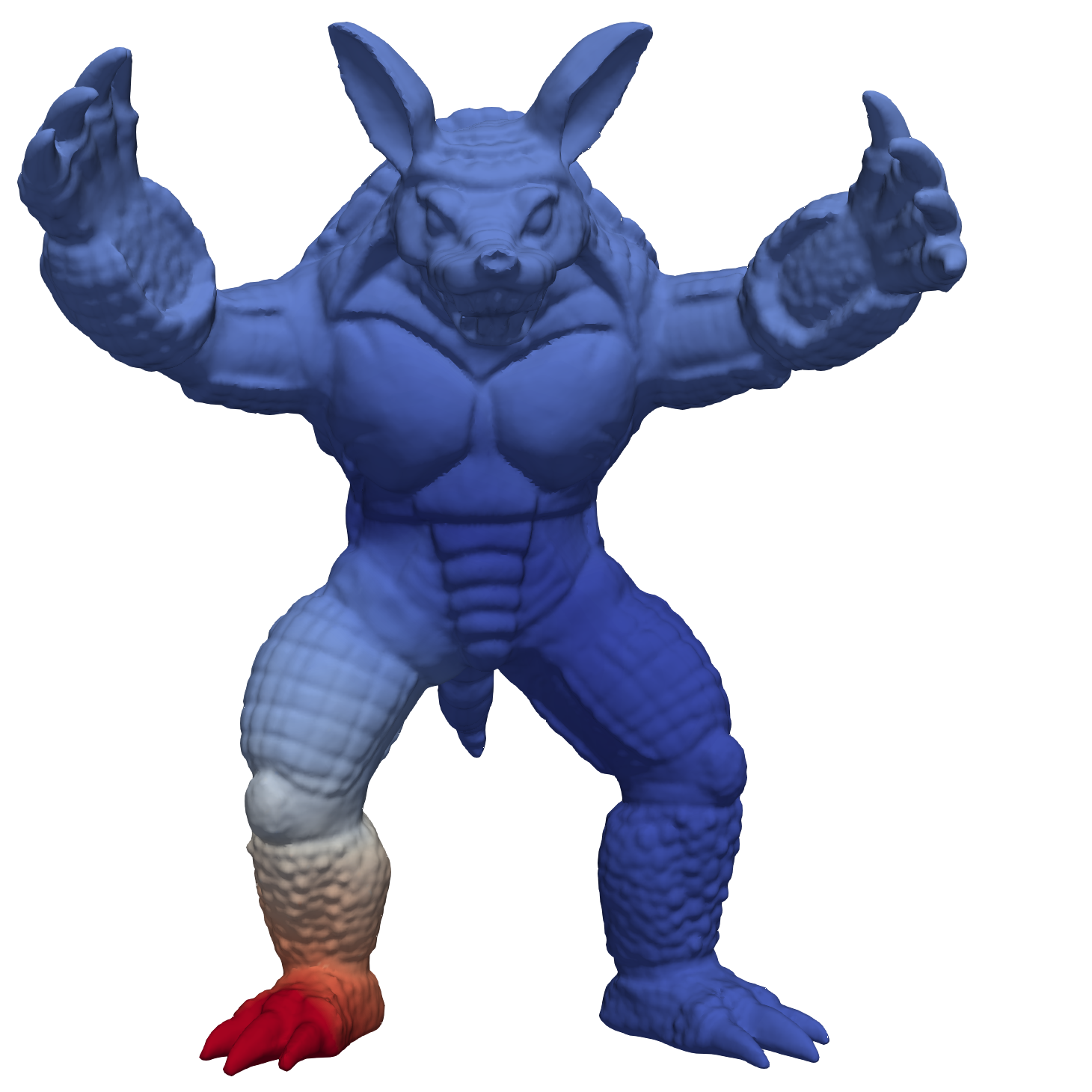}
		\end{subfigure}
		\begin{subfigure}{0.23\linewidth}
			\centering
			\includegraphics[width=\linewidth]{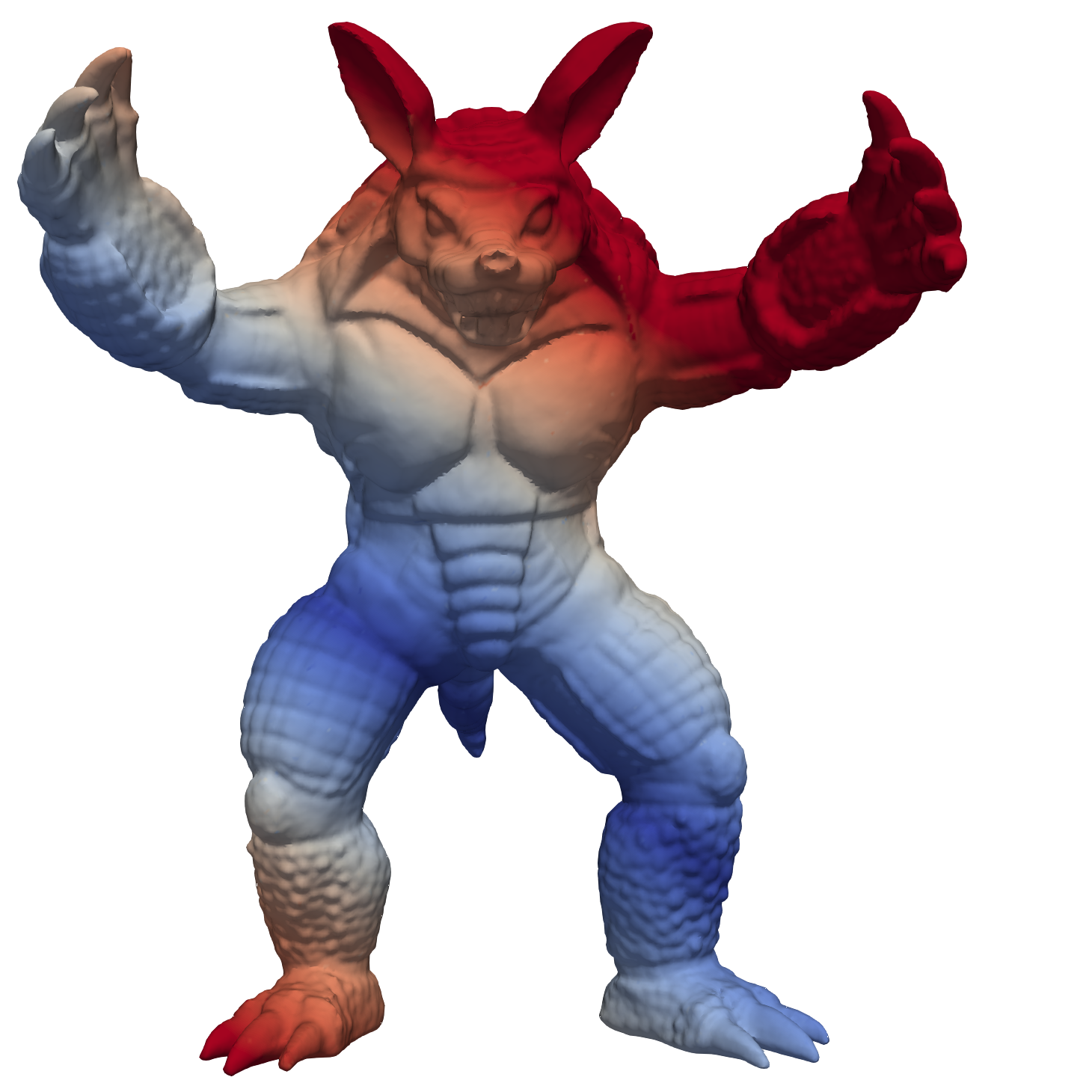}
		\end{subfigure}
	    \\\rotatebox[origin=c]{90}{\makebox[\ImageHt]{Pose 2}}\hspace{0.1em}
		\begin{subfigure}{0.23\linewidth}
			\centering
			\includegraphics[width=\linewidth]{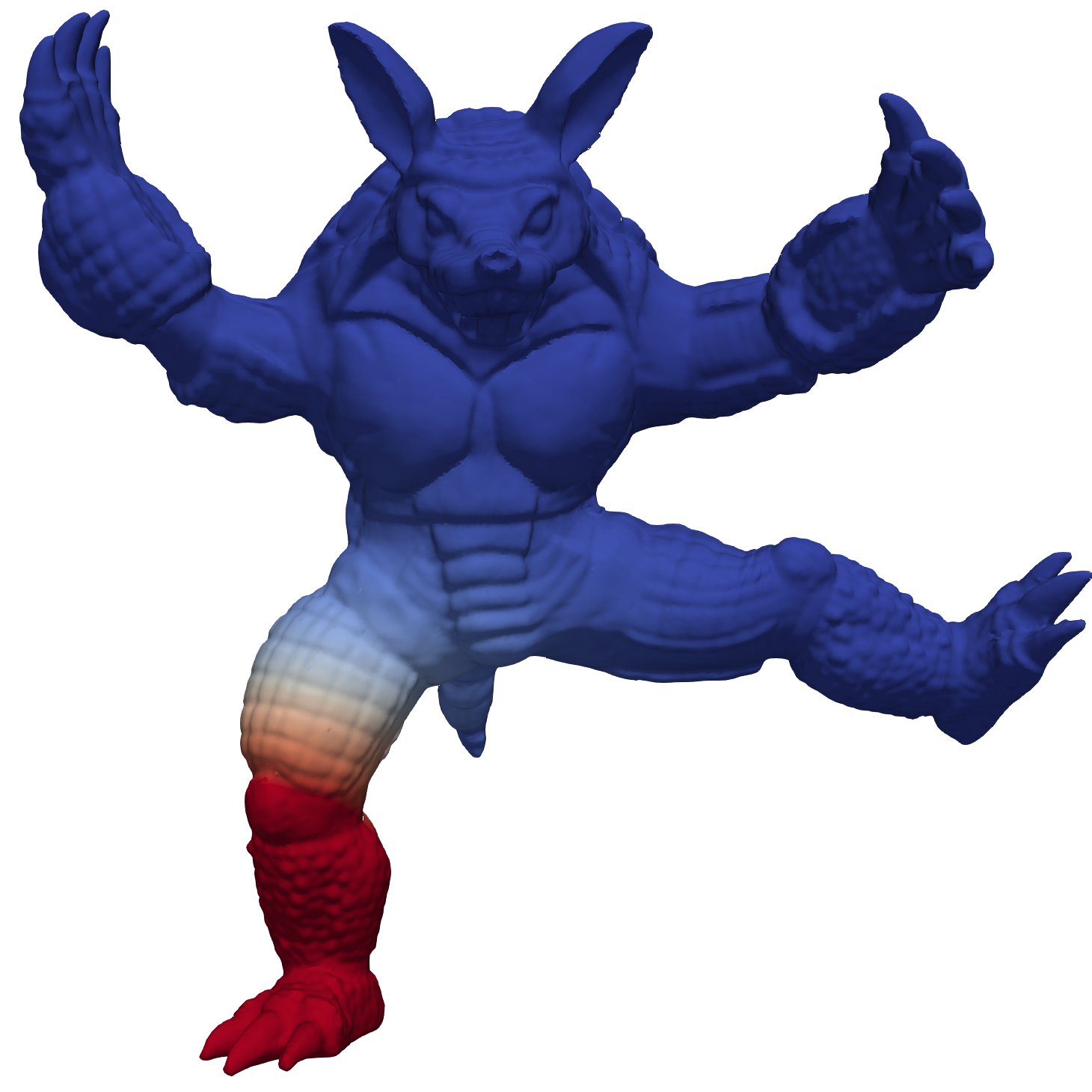}
			\subcaption{}
		\end{subfigure}
		\begin{subfigure}{0.23\linewidth}
        	\centering
			\includegraphics[width=\linewidth]{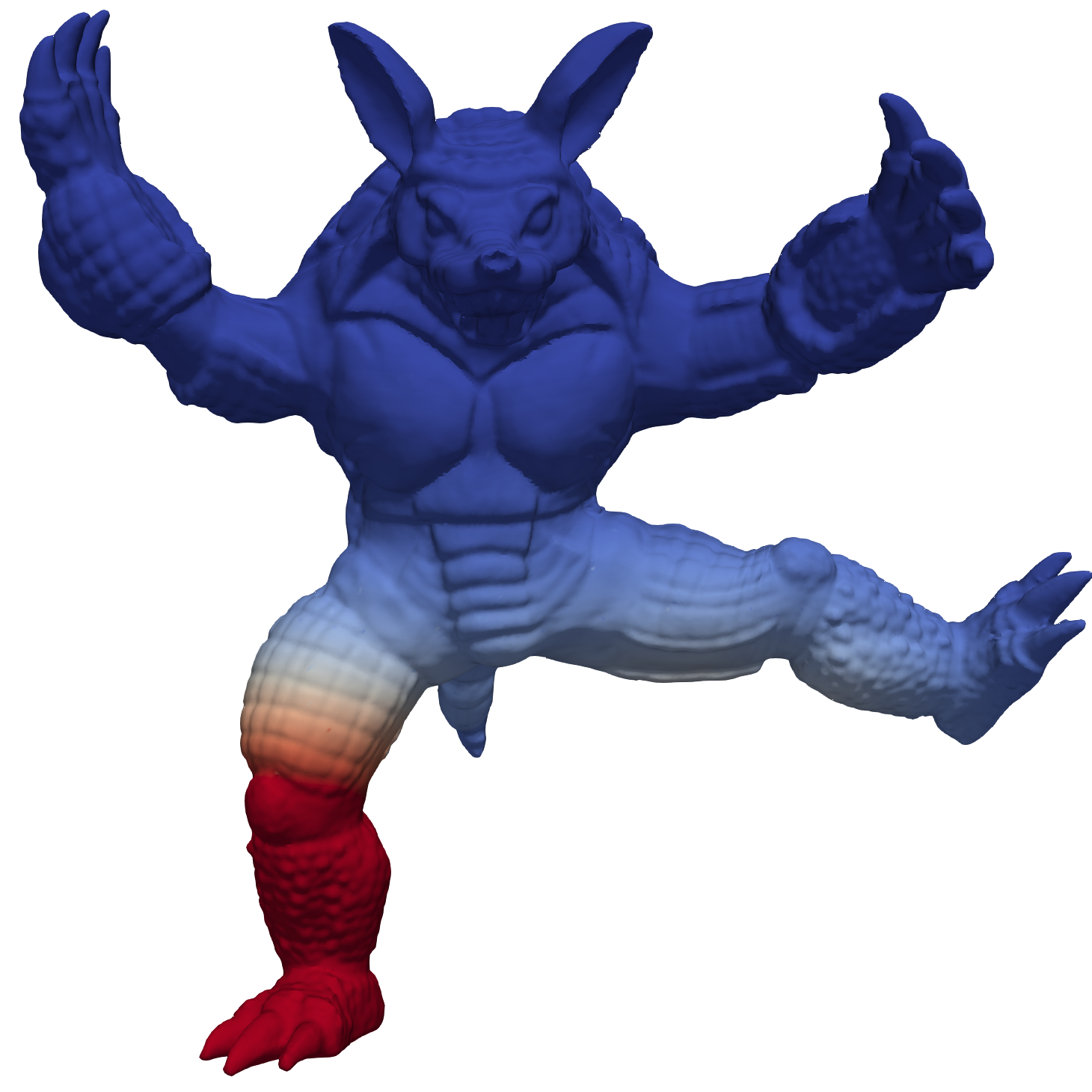}
			\subcaption{}
		\end{subfigure}
		\begin{subfigure}{0.23\linewidth}
			\centering
			\includegraphics[width=\linewidth]{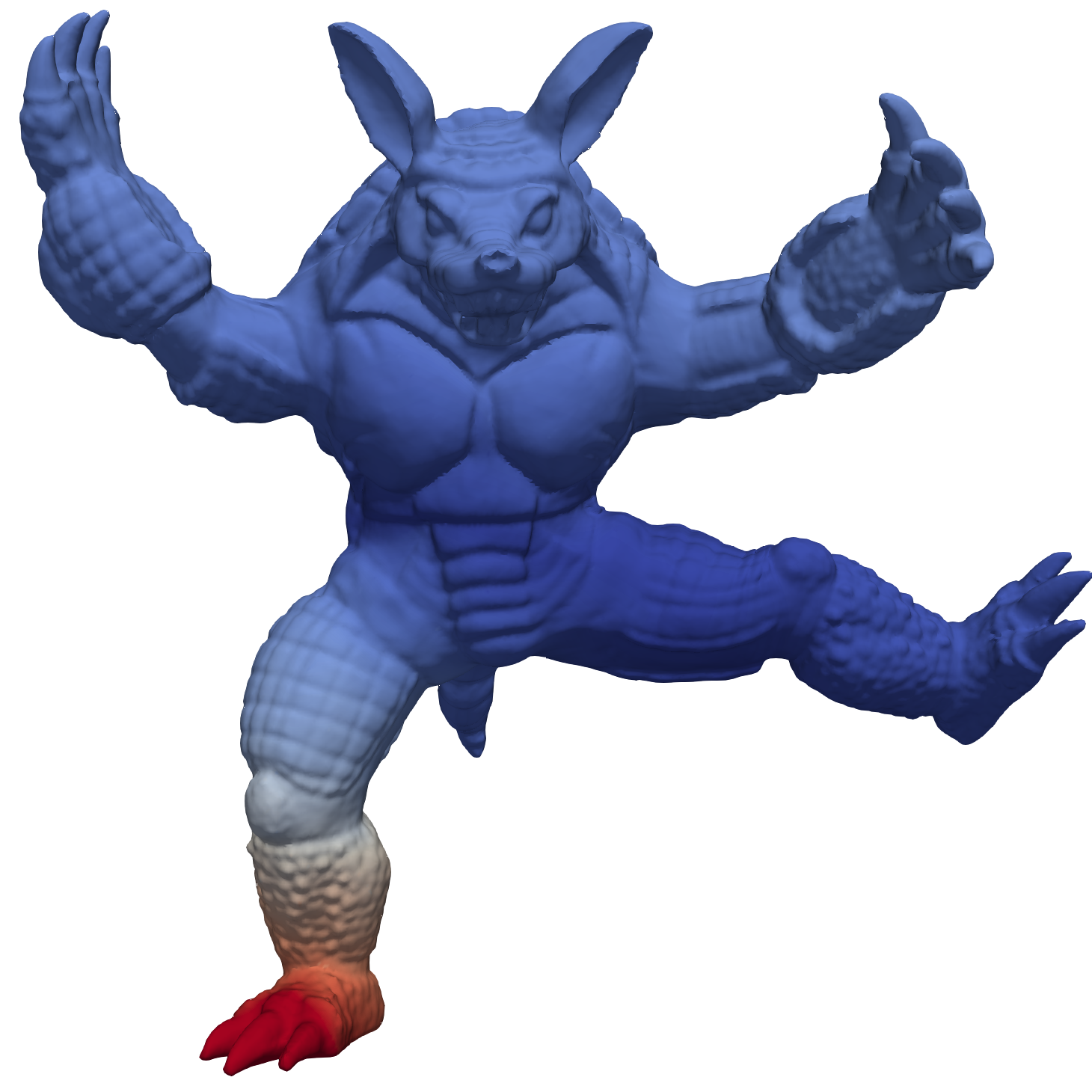}
			\subcaption{}
		\end{subfigure}
		\begin{subfigure}{0.23\linewidth}
			\centering
			\includegraphics[width=\linewidth]{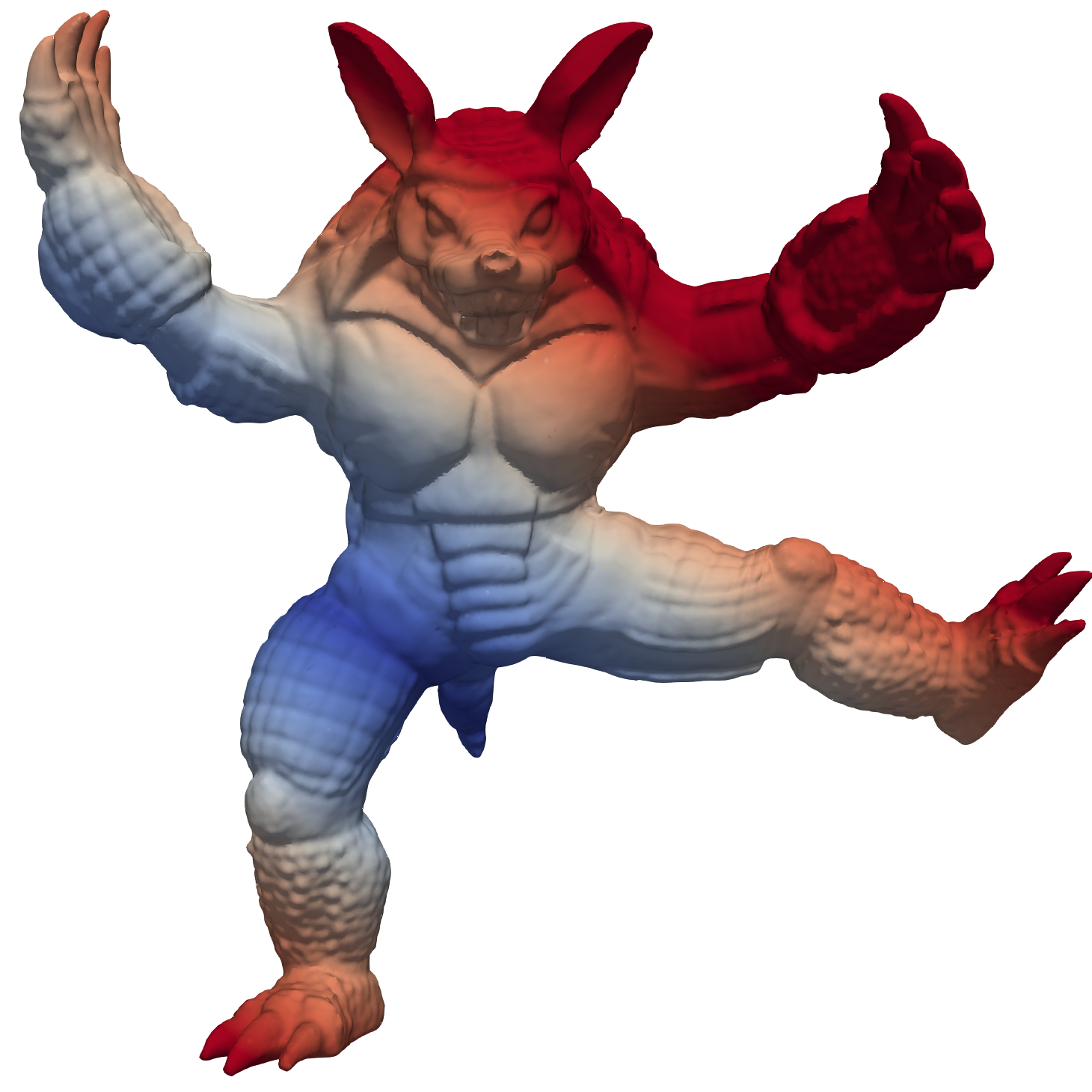}
			\subcaption{}
		\end{subfigure}
	\caption{Visualization of the absolute point correlation to a landmark on the right foot: (a) Gaussian, (b) Symmetrical Gaussian, (c) Generalized inverse Laplacian, (d) Generalized inverse Laplacian and dot product. The point correlation is recomputed based on 2 different poses. For the Gaussian kernels, the correlation changes based on the pose, for the inverse Laplacian kernel, the correlation stays the same, whereas the dot product kernel also adds some correlation 180\textdegree~away from the landmark. Red: High absolute correlation, blue: Low correlation.}
	\label{fig:kernel-distance-to-point}
\end{figure}

\subsubsection{Generalized Inverse Laplacian Correction Term}\label{subsub:invlap-correction}
We inspect the correction term by isolating $Z$ from
\begin{equation}
	(L+I)^{-1}\hat{U}=Z+L^\dagger(L^\dagger+I)^{-1}\hat{U}
\end{equation}
when $W=I$ then $Z$ is exactly the mean observation $\bar{u}$ which is missing from $\hat{X}_R$. The same setup for ICP-A is likewise the average affine transformation that is missing. In practical terms, this means that using the generalized inverse Laplacian in GiNGR cannot correct for a rigid offset like the ICP-A algorithm. One way to overcome this is to additionally add the optimal rigid transformation based on the obtained correspondence pairs. The transformation can be computed in closed form using a least-squares approximate solution \cite{umeyama_least-squares_1991}. Note that the BCPD algorithm uses the same method to separate the rigid transformation from the non-rigid deformations. 

\subsection{ICP and CPD Comparison in GiNGR}\label{sub:icp-cpd-in-giner}
There are two main points that distinguish the ICP and the CPD methods, namely the œkernels that are used and the way that the observed deformations $\hat{U}$ are found. In \cref{fig:kernel-distance-to-point}, the different kernels are visualized, with the color indicating the absolute kernel correlation to a landmark on the right foot of the Armadillo\footnote{\href{http://graphics.stanford.edu/data/3Dscanrep/}{graphics.stanford.edu/data/3Dscanrep/}}. The generalized inverse Laplacian matrix as used in the ICP methods has a direct relation to the commute distance matrix \cite{von_luxburg_tutorial_2007} which measures the distance on the surface. In comparison, the Gaussian kernel operates with the Euclidean distance. A surface based kernel might be favorable to use if the shape that is being analyzed has spatially close but uncorrelated items as for instance the feet of the Armadillo. An interesting observation is how the dot product kernel adds correlation to points located 180\textdegree~away. This helps explain how the ICP-A algorithm works better on the partial data registration experiment in \cite{amberg_optimal_2007}, in comparison to the ICP-T algorithm. In \cref{fig:kernel-distance-to-point} we also included the simple symmetrical kernel from \cite{gerig_morphable_2018} with a symmetrized Gaussian kernel. The symmetrical kernel is useful for registering symmetrical objects such as the face or the skull.
\begin{figure}[t]
	\centering
		\begin{subfigure}{0.325\linewidth}
		\includegraphics[width=\linewidth]{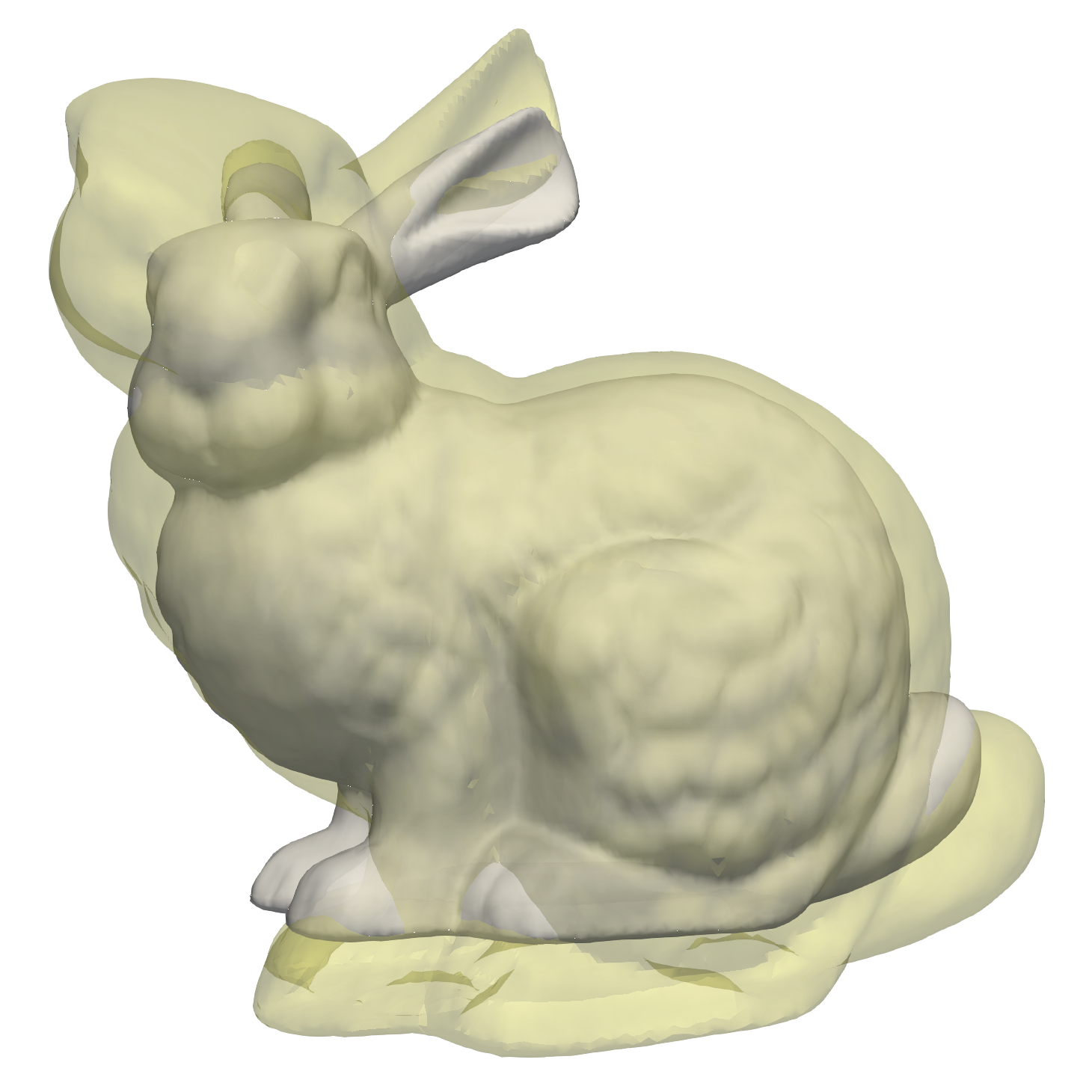}
			\subcaption{}
		\end{subfigure}
		\begin{subfigure}{0.325\linewidth}
			\includegraphics[width=\linewidth]{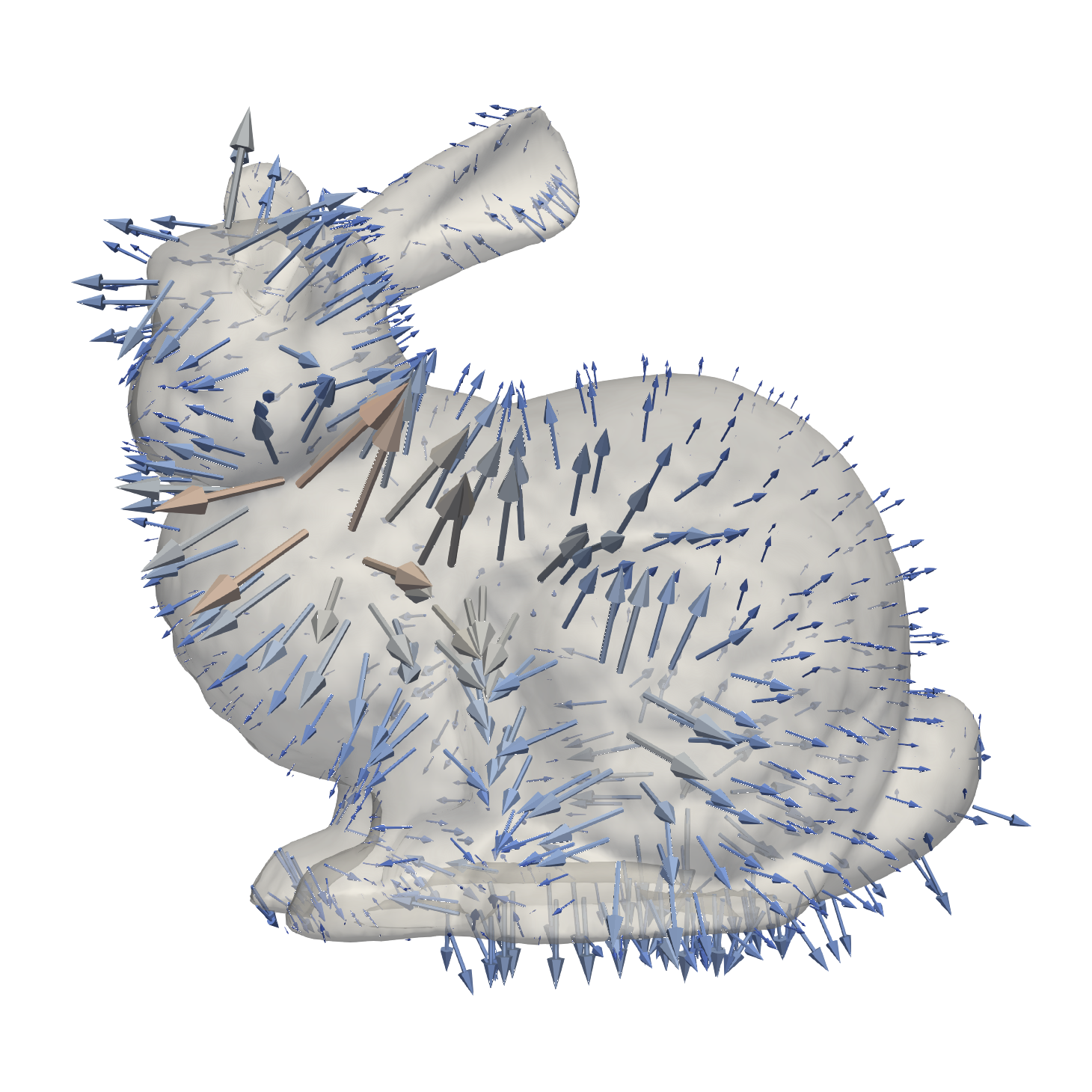}
			\subcaption{}
		\end{subfigure}
		\begin{subfigure}{0.325\linewidth}
			\includegraphics[width=\linewidth]{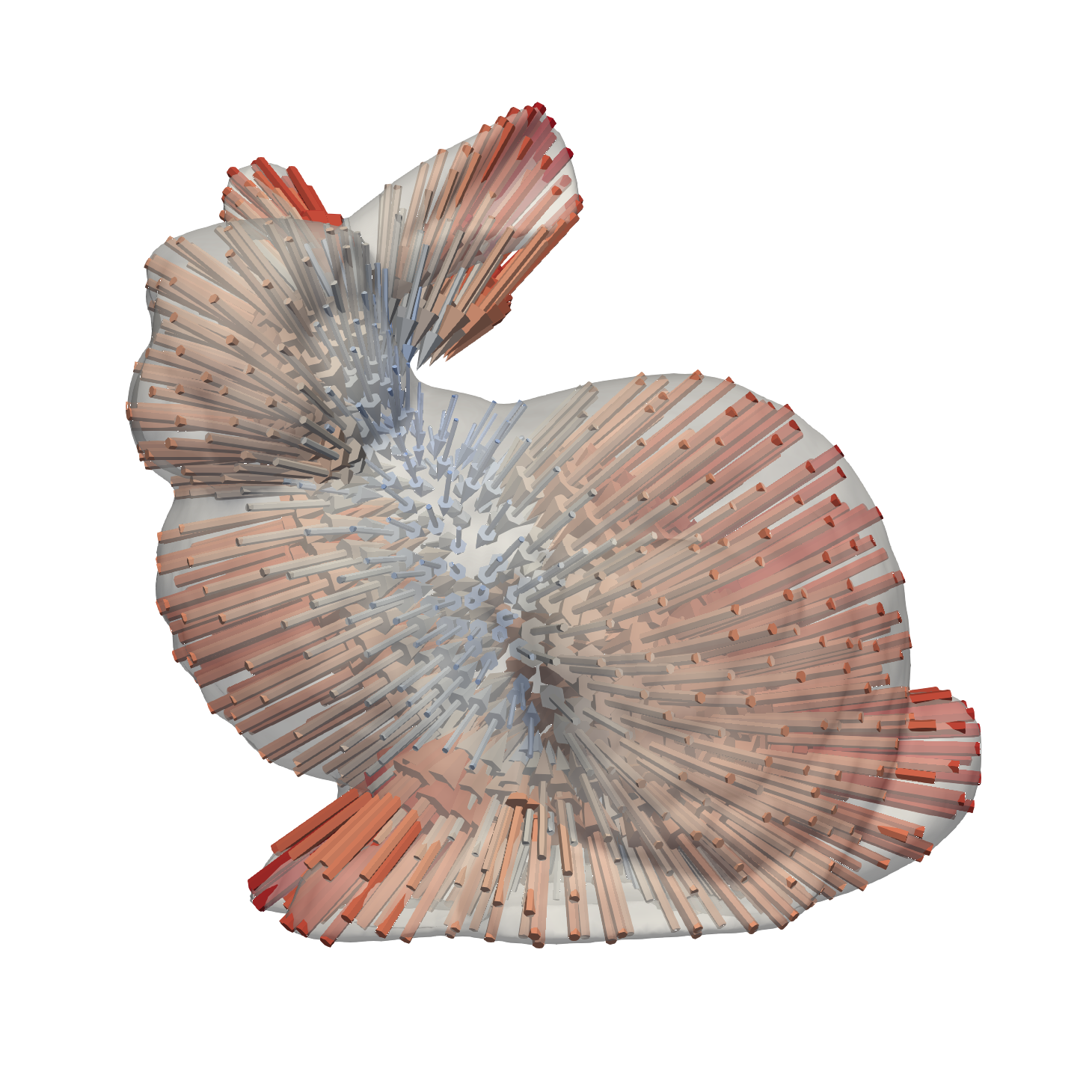}
			\subcaption{}
		\end{subfigure}
	\caption{(a) Reference mesh in white and transparent target mesh in yellow. (b) reference and correspondence pairs from ICP. (c) reference and correspondence pairs from CPD. Notice how the CPD correspondences are not located on the target surface.}
	\label{fig:gingr-correspondence-comparison}
\end{figure}
In \cref{fig:gingr-correspondence-comparison}, we show how the correspondence pairs look for the first iteration of the CPD and the ICP algorithms. The observed deformations in the ICP methods use the closest point heuristic and filter away points based on their normal direction, if a point is on the boundary of a mesh or if the closest point deformation vector ends up intersecting the mesh itself. On the contrary, CPD based methods span a complete correspondence probability matrix between all point pairs in $X_R$ and $X_T$. As can be observed in \cref{fig:gingr-correspondence-comparison}, the probabilistic approach has the effect that corresponding points are initially not found on the target surface. Probabilistic correspondence works well when the global structure of the target is given and can recover large rigid transformation offsets. Another advantage of probabilistic correspondence methods is for noisy point clouds where an ICP approach might have difficulties filtering away noisy points. 

\subsubsection{Deterministic vs Probabilistic GiNGR}\label{subsub:deterministic-vs-probabilistic}
\begin{figure}[t]
	\centering
		\begin{subfigure}{0.19\linewidth}
        	\centering
			\includegraphics[trim={565px 0 565px 0},clip, angle=-11, width=\linewidth]{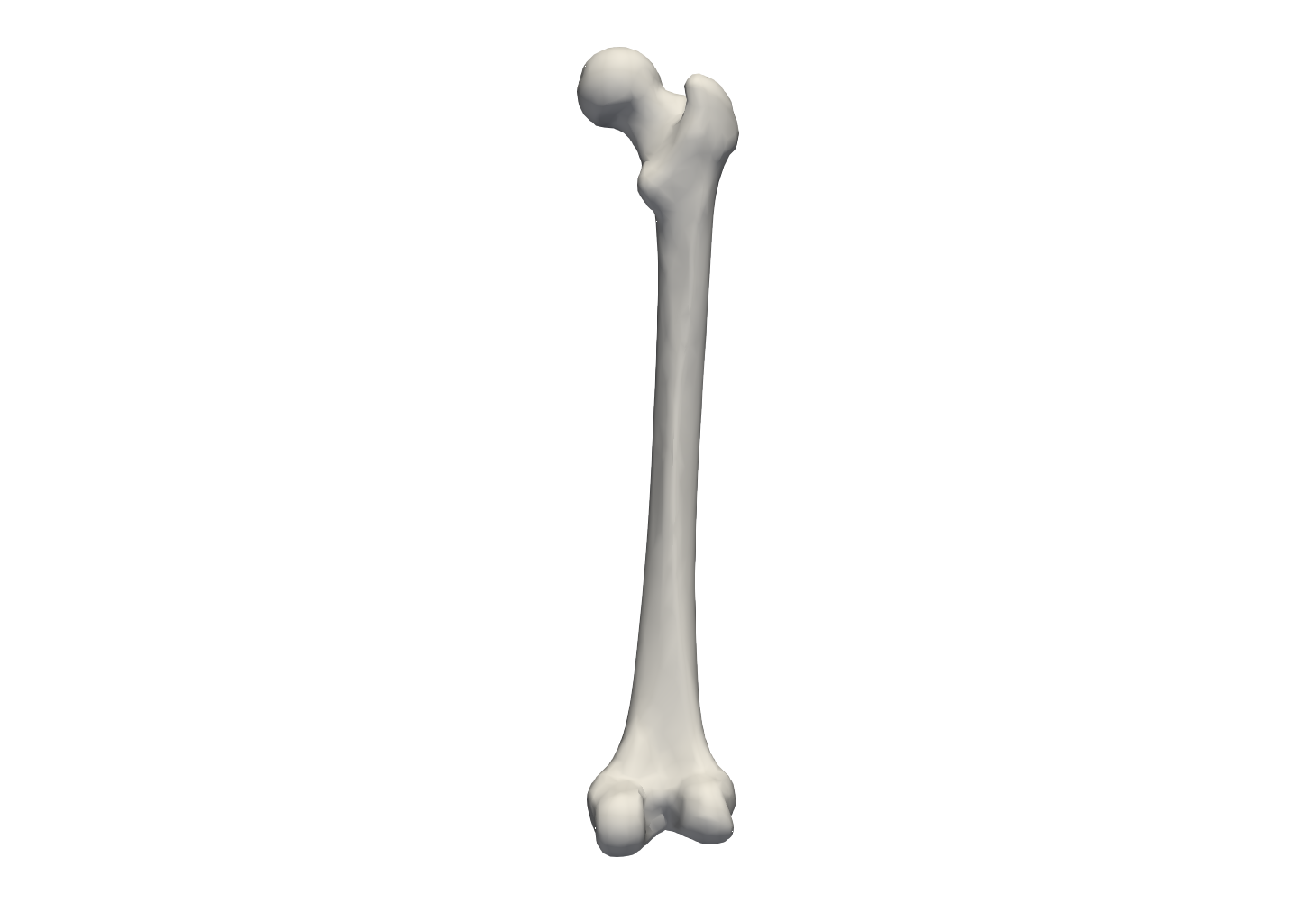}
			\subcaption{}
		\end{subfigure}
		\begin{subfigure}{0.19\linewidth}
			\centering
			\includegraphics[trim={565px 0 565px 0},clip, angle=-11, width=\linewidth]{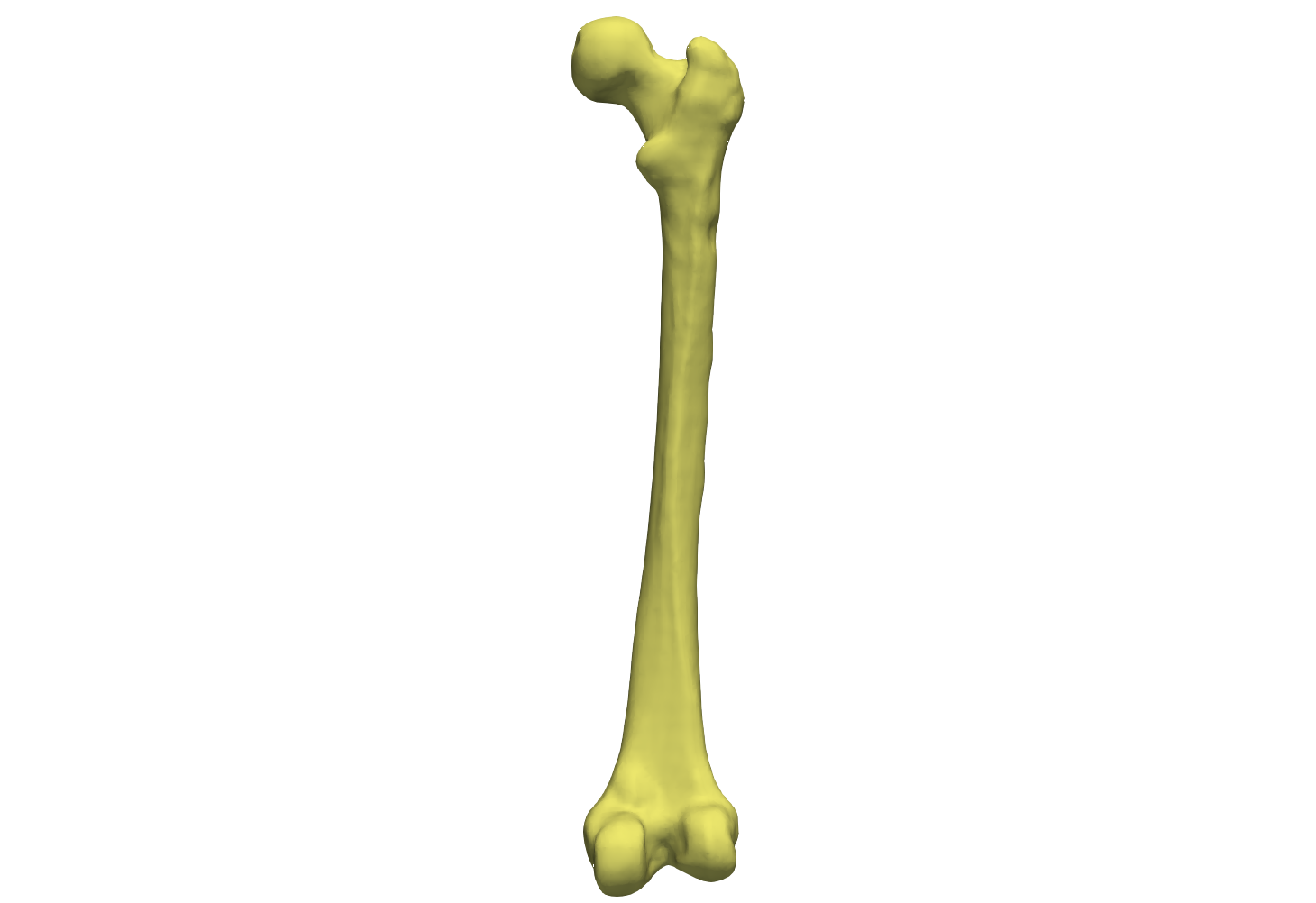}
			\subcaption{}
		\end{subfigure}
		\begin{subfigure}{0.19\linewidth}
			\centering
			\includegraphics[trim={565px 0 565px 0},clip, angle=-11, width=\linewidth]{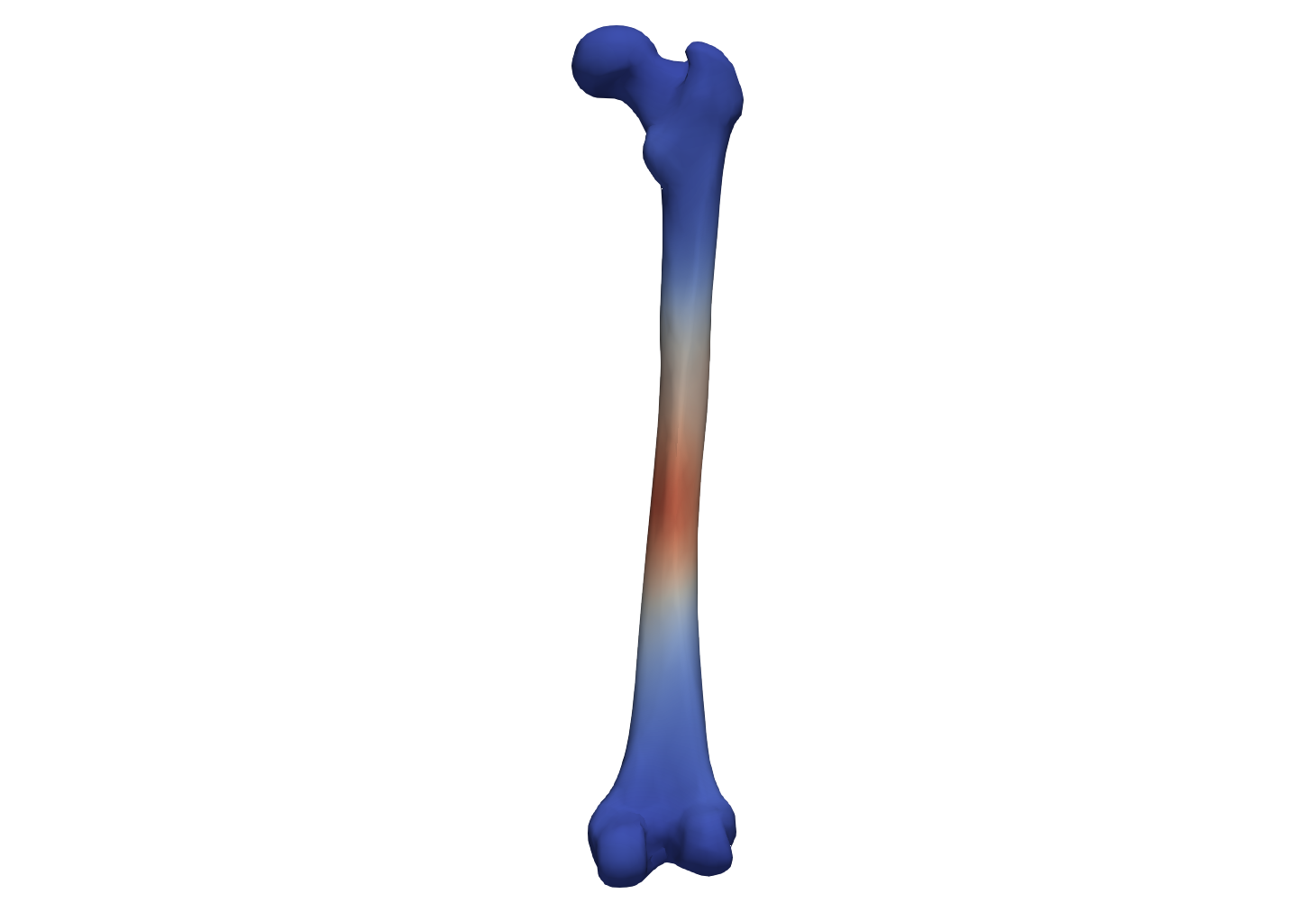}
			\subcaption{}
		\end{subfigure}
		\begin{subfigure}{0.19\linewidth}
			\centering
			\includegraphics[trim={565px 0 565px 0},clip, angle=-11, width=\linewidth]{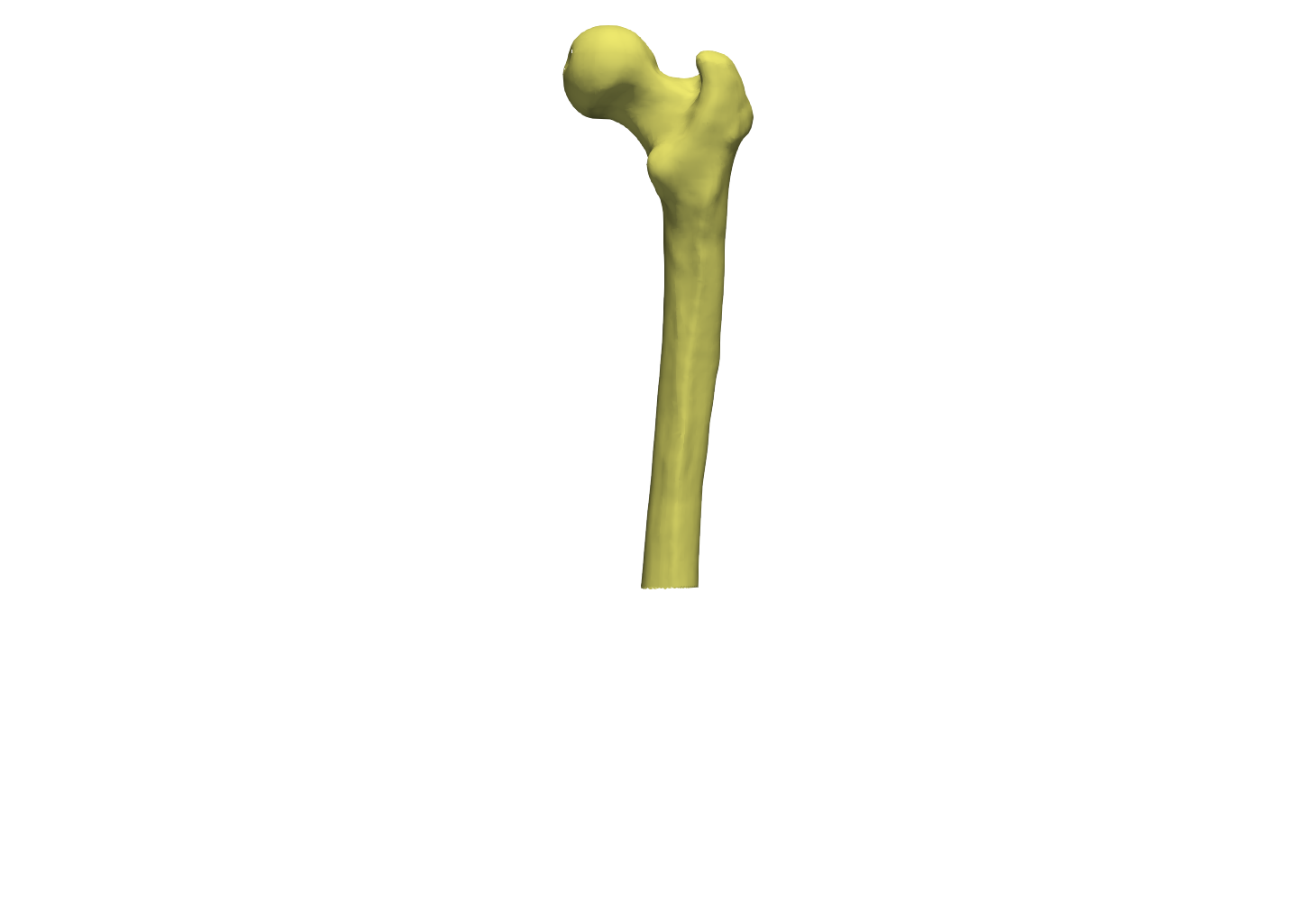}
			\subcaption{}
		\end{subfigure}
		\begin{subfigure}{0.19\linewidth}
			\centering
			\includegraphics[trim={565px 0 565px 0},clip, angle=-11, width=\linewidth]{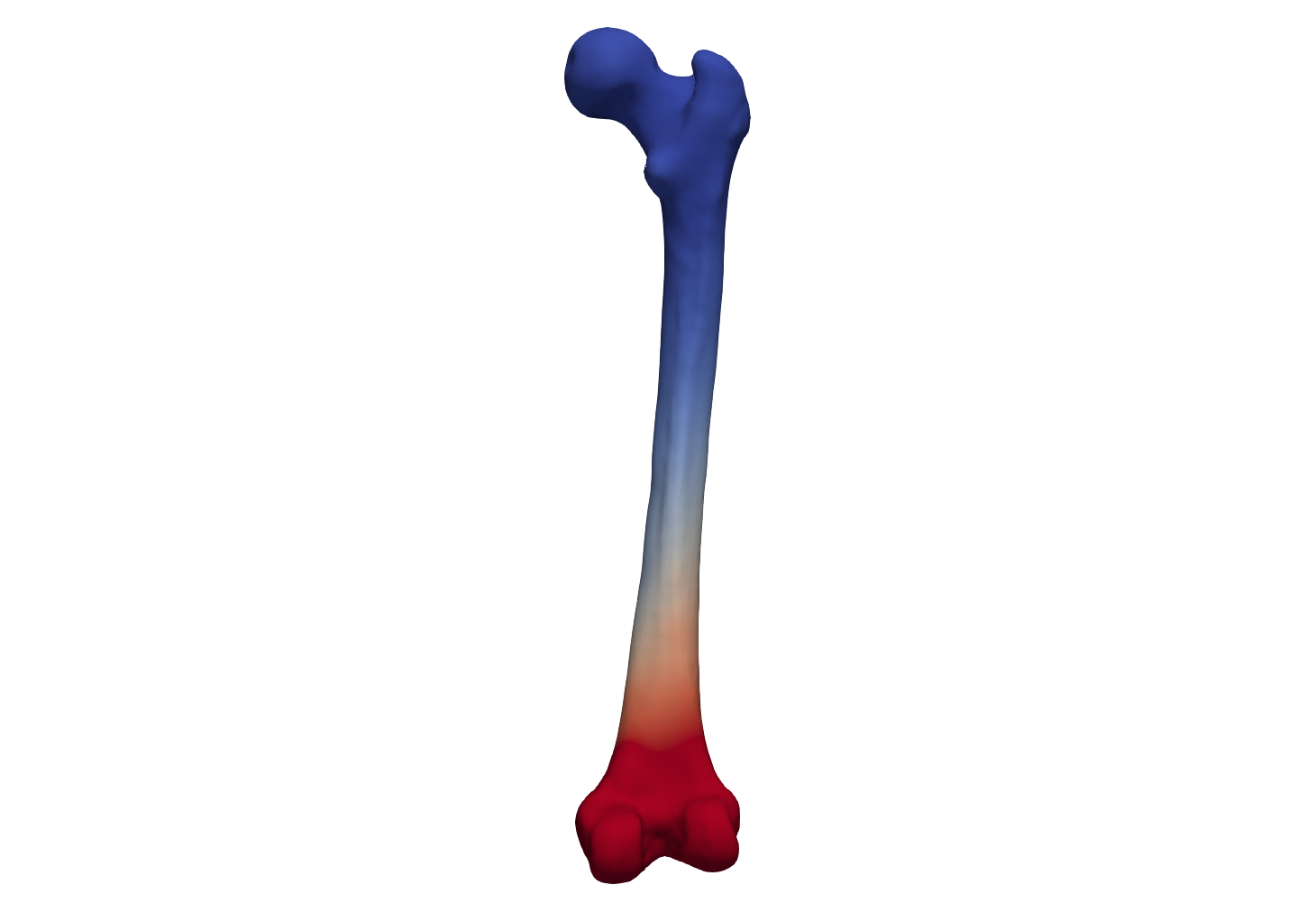}
			\subcaption{}
		\end{subfigure}
	\caption{Posterior visualization of registration uncertainty. (a) Reference, (b, c) complete target and visualization of the posterior, (d, e) partial target and visualization of the posterior. Red: High absolute correlation, blue: Low correlation.}
	\label{fig:posterior-visualization}
\end{figure}

\begin{figure}[t]
	\centering
		\begin{subfigure}{0.49\linewidth}
			\includegraphics[width=\linewidth]{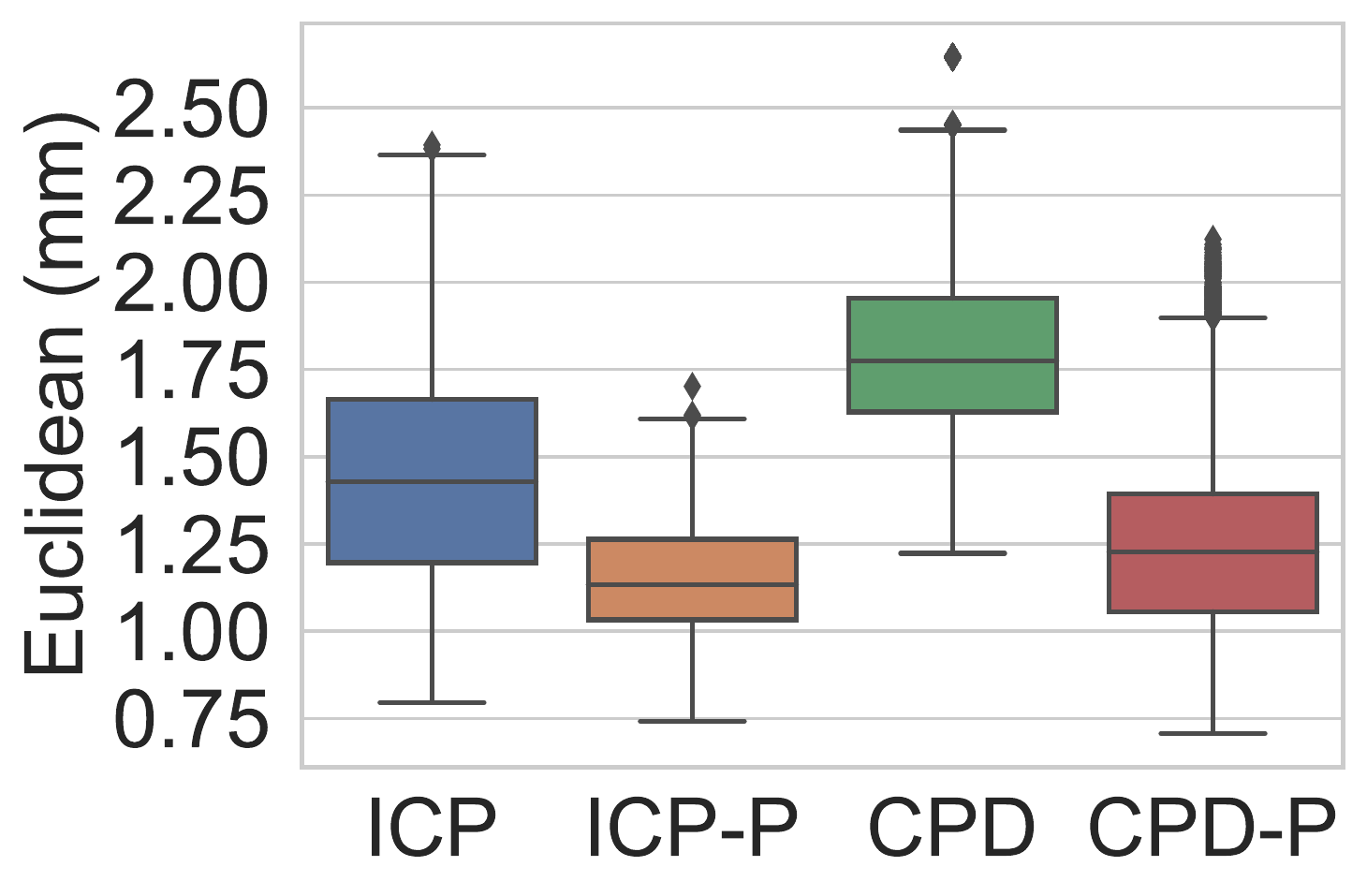}
			\subcaption{}
		\end{subfigure}
		\begin{subfigure}{0.49\linewidth}
			\includegraphics[width=\linewidth]{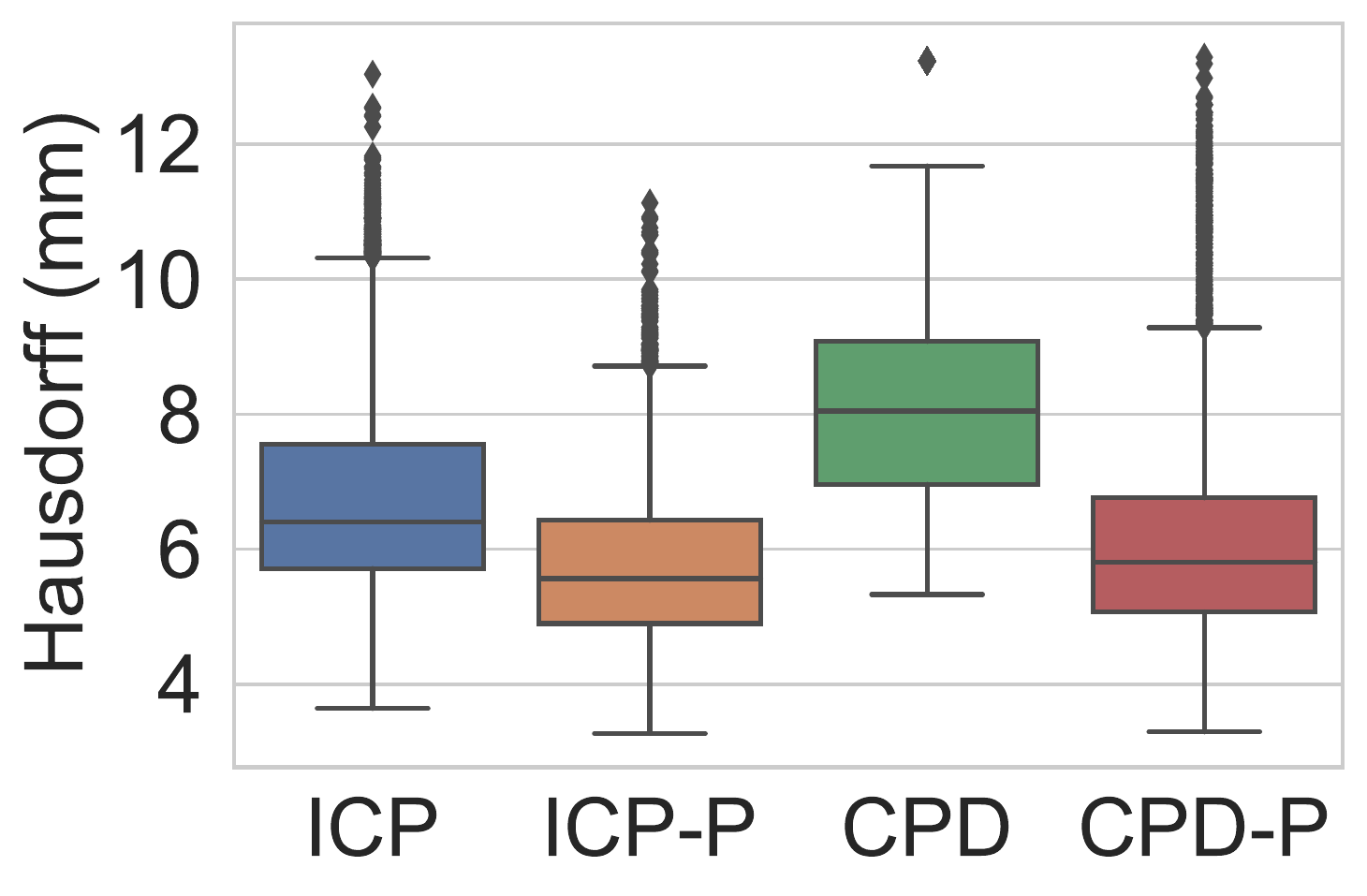}
			\subcaption{}
		\end{subfigure}
	\caption{Comparison of registration performance between ICP and CPD in either deterministic or probabilistic (-P) mode. Both the average Euclidean and Hausdorff distances are shown from 5000 registrations of femurs.}
	\label{fig:gingr-best-registration-comparison}
\end{figure}

Besides the ability to compare existing algorithms based on their kernel and correspondence function in GiNGR, existing registration algorithms can also be converted to probabilistic methods in comparison to their deterministic nature. With probabilistic registration one is able to estimate the posterior distribution of possible correspondence pairs. In \cref{fig:posterior-visualization}, we show an example of registering a complete and a partial femur bone and how the posterior can be used to visualize the uncertainty in the obtained correspondence pairs. Note how the complete femur even has a large uncertainty along the shaft due to missing anatomical stable points.
Probabilistic registration also has the potential to improve the single best registration accuracy. In \cref{fig:gingr-best-registration-comparison}, we compare ICP and CPD registrations over 5000 different reference and target femur pairs from \cite{madsen_closest_2020} using both the deterministic and the probabilistic settings of GiNGR. For this experiment, the femurs are first aligned based on 6 landmarks, before the registration begins. The final result shows that the probabilistic setting on average is better than the deterministic default for this particular dataset.

\subsection{Point Distribution Models (PDMs)}\label{sub:pdm}
GiNGR can be used with analytical kernels to bring point sets in correspondence to build a statistical kernel. This makes the modeled deformations equivalent to a PDM \cite{cootes_training_1992, luthi_gaussian_2017}. With a statistical kernel, we can use GiNGR the same way as with analytical kernels. The specific case of using ICP to fit a PDM has also previously been shown \cite{cheng_statistical_2017}.

\subsection{Expert annotations \& Multi-resolution}\label{sub:multires}
In \cite{amberg_optimal_2007}, they show how expert annotated landmarks can be integrated in the optimization and controlled via an extra hyperparameter. In contrast, CPD and BCPD does not allow for expert annotations. With algorithms formulated in GiNGR, the inclusion of expert annotations is however trivial as they can be manually added to the set of correspondence pairs as noisy observations to the GPR formulation as also explained in \cite{morel-forster_probabilistic_2018}. 

GiNGR also allows for multi-resolution registration which can be useful for time-critical applications. It works by performing the registration on down-sampled instances. After the registration, the full-resolution GPMM is used to interpolate the high-resolution details. The same principle is also used in \cite{hirose_acceleration_2020} to speed up the BCPD algorithm.

\section{Related Work}\label{sec:relatedwork}
In \cref{sub:icp_cpd} we showed how the CPD and BCPD can be reformulated to GiNGR. Furthermore, we showed how the least-squares ICP implementations can be approximated into GiNGR using the Woodbury matrix identity. In this section, we highlight how other non-rigid registration methods estimate the correspondence field $\hat{U}$ and what kind of smoothness regularization they apply.

\subsection{Correspondence Estimation}
In \cite{chui_new_2003}, a binary assignment method to establish correspondence between two point sets is introduced. A sparse correspondence matrix is used to identify corresponding points and potential outliers. The method also allows for soft-correspondence assigning values, which is similar to the $P$ matrix in the CPD method, but without explicitly formulating the registration as a maximum likelihood estimate.

In \cite{chetverikov_robust_2005} they introduce the Trimmed ICP (TrICP), which sorts the closest points according to their Euclidean distance to the target. The closest points below a set distance threshold are filtered to only compute the transformation on a subset of the points. In \cite{mcneill_probabilistic_2006}, they introduce a probabilistic point matching (PPM), which is used in CPD. Later, they extend their work to part-based PPM \cite{mcneill_part-based_2006} which allows having local different transformations.

In \cite{hufnagel_generation_2008} they generalize the rigid EM-ICP \cite{granger_multi-scale_2002} to affine transformations. The EM-ICP assumes Gaussian noise on the target points and uses an EM-like algorithm to maximize the correspondence probability of the reference points. They also show that it tends toward the standard ICP when a small variance is chosen. This is again a probabilistic correspondence method for point sets that is robust to noise and works similar to CPD. 

Several feature-based alternatives exist to probabilistic correspondence estimation and the closest point heuristic. The spin image is used as a feature descriptor for 3D surfaces \cite{johnson_using_1999} as used in \cite{ma_robust_2015} or fast point feature histogram (FPFH) \cite{rusu_fast_2009} as used in \cite{ma_non-rigid_2016, ma_nonrigid_2019} which captures the underlying surface model properties (local geometry, surface normals, curvatures etc.). A three-stage iterative point registration process is used in \cite{feldmar_rigid_1996} where the last step involves local affine transformations for each vertex. A closest point method based on 8- dimensional (vertex position, vertex normals, and curvature) is used to estimate the correspondence between the two surfaces. A multi-level registration is used in \cite{brown_global_2007}, where they first use feature points as the correspondence estimate and in a later stage change to locally weighted ICP in order to register local details. In \cite{liang_nonrigid_2018}, they increase the reliability of simple closest point correspondence estimation, by introducing a 2-way closest point search. This simple filtering step can be an effective heuristic to remove bad correspondence estimates.

All of the mentioned methods have the potential to be used in GiNGR as they provide a simple point deformation estimate. This list of correspondence estimation methods is by no means complete, but it provides an overview of the existing methods depending on the application at hand. 

\subsection{Smoothness Regularization}
In \cref{sub:icp-cpd-in-giner}, we showed how the regularization used in CPD and the ICP methods lead to different kernel priors and we visually compared different kernels in \cref{fig:kernel-distance-to-point}. In \cite{steinke_kernels_2008}, they show how the covariance operator can generally be seen as an intermediate representation to convert between kernel functions, RKHS, regularization operators, and GPs. The majority of registration papers describe their optimization and regularization terms in a single algorithm, instead of separating them as in GiNGR by using GPMMs. Besides the kernels already mentioned, in \cite{hirose_bayesian_2020} they show how different positive-definite kernels can be used in the CPD algorithm, such as the inverse multiquadric and the rational quadratic kernel. Another popular regularization prior is thin-plate splines (TPS), as used in \cite{chui_new_2003} to define the non-rigid deformations on a range of synthetic data and brain-structure point sets. TPS is also used in \cite{brown_global_2007} for range scanning registration. In \cite{zou_non-rigid_2007} they use a spherical analog of TPS for inter-subject brain surface registration and in \cite{gerig_morphable_2018} they use a multi-scale B-spline kernel together with a symmetrical kernel to register 3D scans of faces. Unfortunately, to our knowledge, there is no single best kernel that works in all domains. The kernel function needs to be designed on a case-by-case basis. Kernels can also be added or multiplied together to complement each other. A complete list of valid kernel mixtures is given in \cite{duvenaud_automatic_2014}.

\subsection{Gaussian Process Regression (GPR)}
GPR has previously been used in different registration pipelines. In \cite{hitchcox_point_2020} they are working in a SLAM environment and use GPR to extract key-points from underwater sonar images. A simple squared exponential kernel is used to model low-frequency elevation changes on the seafloor. In a fine-tuning step, an ICP like approach is used. In \cite{wang_robust_2017}, GPR is used in an iterative fashion to include more and more points into the training set to estimate the road shape based on multiple 3D lidar scans. Starting from a few inlier points, they iteratively add or reject more points to the inlier dataset based on the GP posterior. This way of iteratively adding robust points to the observations can easily be integrated as a strategy in GiNGR. In \cite{madsen_closest_2020} they are using GPMMs and closest point correspondence to obtain a distribution of possible surface registrations using the MH algorithm. They empirically show how the stochastic method improves a simple ICP algorithm by being able to escape local minima. The method can be seen as an extension to GiNGR for applications that are primarily interested in the distribution of different registrations instead of a single best registration.

\section{Technical Details \& Discussion}
With the introduction of GiNGR we have not invented a new registration algorithm, but instead, a framework to facilitate comparison of existing and development of future registration methods. For comparison of the most common registration methods on different datasets, we refer to \cite{hirose_bayesian_2020}. 

If the generalized inverse Laplacian matrix is used, it is important to know that the complete matrix needs to be spanned before a low-rank approximation can be computed. This is however not different from the ICP-T and ICP-A methods themselves, where the complete matrixes are also spanned, making the methods memory intensive for large datasets. The Gaussian kernel can in comparison be efficiently approximated based on e.g. the Nyström method \cite{luthi_gaussian_2017, hirose_bayesian_2020}. 

For the probabilistic setting of GiNGR, we have adapted the default likelihood function from \cite{morel-forster_probabilistic_2018}. However, the framework allows for easy use of any other likelihood functions, such as the collective or the landmark likelihood \cite{schonborn_markov_2017}, the exponential Hausdorff \cite{madsen_closest_2020} and others. In \cite{madsen_closest_2020} they use a step-length hyperparameter to regularize the deformations and allow for a higher acceptance rate in the MH algorithm. Instead, we mix the informed proposal with an additional random-walk proposal in the model parameter space. Mixing of proposals for MH is explained in \cite{schonborn_markov_2017}. The mixture increases the acceptance rate as the random-walk proposals have a high backward transition probability in comparison to the informed proposal alone which is highly asymmetric. 

\section{Conclusion}
In this paper, we have shown how popular non-rigid registration methods for point-set and surface registration can be generalized with the GiNGR framework. The unifying framework separates the optimization from the modeling and allows for a direct comparison of registration methods. We have shown how ICP algorithms are fundamentally different from popular CPD based algorithms by comparing their regularization, correspondence, and robustness attributes. We have also shown how existing algorithms can be converted to probabilistic registration in contrast to their deterministic nature. By explaining ICP algorithms with GiNGR, we have identified that ICP-A is better in registering partial surfaces with a small rigid offset due to the dot product kernel in comparison to the ICP-T algorithm. In the CPD algorithm, we have identified that the underlying independent noise assumption is individually computed for each point based on the correspondence probability matrix. Beyond comparing existing registration algorithms, GiNGR can be used to easily change between different methods during registration or to create methods that combine parts from existing algorithms, e.g. CPD using a Laplacian-based kernel. The framework gives additional benefits such as a clear concept for including expert annotation, multi-resolution registration for faster and more robust registration procedures, and the ability to use statistical deformation priors. With its modular design and open-source availability, we hope to bring different non-rigid point-set and surface registration communities together.

{\small
\bibliographystyle{splncs04}
\bibliography{egbib}

\begin{thebibliography}{10}
\providecommand{\url}[1]{\texttt{#1}}
\providecommand{\urlprefix}{URL }
\providecommand{\doi}[1]{https://doi.org/#1}

\bibitem{allen_space_2003}
Allen, B., Curless, B., Popovic, Z.: The space of human body shapes:
  reconstruction and parameterization from range scans. TOG  \textbf{22}(3),
  587--594 (2003)

\bibitem{amberg_optimal_2007}
Amberg, B., Romdhani, S., Vetter, T.: Optimal step nonrigid {ICP} algorithms
  for surface registration. In: 2007 {IEEE} {Conference} on {Computer} {Vision}
  and {Pattern} {Recognition}. pp.~1--8. IEEE (2007)

\bibitem{besl_method_1992}
Besl, P.J., McKay, N.D.: A method for registration of 3-{D} shapes. IEEE
  Transactions on Pattern Analysis and Machine Intelligence  \textbf{14}(2),
  239--256 (Feb 1992). \doi{10.1109/34.121791}

\bibitem{bogo_faust_2014}
Bogo, F., Romero, J., Loper, M., Black, M.J.: {FAUST}: {Dataset} and evaluation
  for {3D} mesh registration. In: Proceedings of the {IEEE} {Conference} on
  {Computer} {Vision} and {Pattern} {Recognition}. pp. 3794--3801 (2014)

\bibitem{brown_global_2007}
Brown, B., Rusinkiewicz, S.: Global non-rigid alignment of 3-d scans. ACM
  Trans. Graph.  \textbf{26}, ~21 (07 2007). \doi{10.1145/1275808.1276404}

\bibitem{chen_object_1992}
Chen, Y., Medioni, G.: Object modelling by registration of multiple range
  images. Image and vision computing  \textbf{10}(3),  145--155 (1992)

\bibitem{cheng_statistical_2017}
Cheng, S., Marras, I., Zafeiriou, S., Pantic, M.: Statistical non-rigid {ICP}
  algorithm and its application to {3D} face alignment. Image and Vision
  Computing  \textbf{58},  3--12 (Feb 2017). \doi{10.1016/j.imavis.2016.10.007}

\bibitem{chetverikov_robust_2005}
Chetverikov, D., Stepanov, D., Krsek, P.: Robust {Euclidean} alignment of {3D}
  point sets: the trimmed iterative closest point algorithm. Image and Vision
  Computing  \textbf{23}(3),  299--309 (2005).
  \doi{10.1016/j.imavis.2004.05.007}

\bibitem{chui_new_2003}
Chui, H., Rangarajan, A.: A new point matching algorithm for non-rigid
  registration. Computer Vision and Image Understanding  \textbf{89}(2),
  114--141 (Feb 2003). \doi{10.1016/S1077-3142(03)00009-2}

\bibitem{cootes_training_1992}
Cootes, T.F., Taylor, C.J., Cooper, D.H., Graham, J.: Training models of shape
  from sets of examples. In: {BMVC92}, pp. 9--18. Springer (1992)

\bibitem{duvenaud_automatic_2014}
Duvenaud, D.: Automatic model construction with {Gaussian} processes. {PhD}
  {Thesis}, University of Cambridge (2014)

\bibitem{feldmar_rigid_1996}
Feldmar, J., Ayache, N.: Rigid, affine and locally affine registration of
  free-form surfaces. Int J Comput Vision  \textbf{18}(2),  99--119 (May 1996).
  \doi{10.1007/BF00054998}

\bibitem{gerig_morphable_2018}
Gerig, T., Morel-Forster, A., Blumer, C., Egger, B., Luthi, M., Sch{\"o}nborn,
  S., Vetter, T.: Morphable face models-an open framework. In: 2018 13th {IEEE}
  {International} {Conference} on {Automatic} {Face} \& {Gesture} {Recognition}
  ({FG} 2018). pp. 75--82. IEEE (2018)

\bibitem{granger_multi-scale_2002}
Granger, S., Pennec, X.: Multi-scale {EM}-{ICP}: {A} fast and robust approach
  for surface registration. In: European {Conference} on {Computer} {Vision}.
  pp. 418--432. Springer (2002)

\bibitem{gutman_generalized_2004}
Gutman, I., Xiao, W.: Generalized inverse of the {Laplacian} matrix and some
  applications. Bull Class Sci Math  \textbf{129}(29),  15--23 (2004).
  \doi{10.2298/BMAT0429015G}

\bibitem{hirose_acceleration_2020}
Hirose, O.: Acceleration of non-rigid point set registration with downsampling
  and {Gaussian} process regression. IEEE Transactions on Pattern Analysis and
  Machine Intelligence pp.~1--1 (2020). \doi{10.1109/TPAMI.2020.3043769}

\bibitem{hirose_bayesian_2020}
Hirose, O.: A {Bayesian} {Formulation} of {Coherent} {Point} {Drift}. IEEE
  Transactions on Pattern Analysis and Machine Intelligence pp.~1--1 (2020).
  \doi{10.1109/TPAMI.2020.2971687}

\bibitem{hitchcox_point_2020}
{Hitchcox}, T., {Forbes}, J.R.: A point cloud registration pipeline using
  gaussian process regression for bathymetric slam*. In: 2020 IEEE/RSJ
  International Conference on Intelligent Robots and Systems (IROS). pp.
  4615--4622 (2020). \doi{10.1109/IROS45743.2020.9340944}

\bibitem{hontani_robust_2012}
Hontani, H., Matsuno, T., Sawada, Y.: Robust nonrigid {ICP} using
  outlier-sparsity regularization. In: 2012 {IEEE} {Conference} on {Computer}
  {Vision} and {Pattern} {Recognition}. pp. 174--181. IEEE (2012)

\bibitem{hufnagel_generation_2008}
Hufnagel, H., Pennec, X., Ehrhardt, J., Ayache, N., Handels, H.: Generation of
  a statistical shape model with probabilistic point correspondences and the
  expectation maximization- iterative closest point algorithm. Int J CARS
  \textbf{2}(5),  265--273 (2008). \doi{10.1007/s11548-007-0138-9}

\bibitem{jiang_disentangled_2020}
Jiang, B., Zhang, J., Cai, J., Zheng, J.: Disentangled {Human} {Body}
  {Embedding} {Based} on {Deep} {Hierarchical} {Neural} {Network}. IEEE
  Transactions on Visualization and Computer Graphics  \textbf{26}(8),
  2560--2575 (Aug 2020). \doi{10.1109/TVCG.2020.2988476}

\bibitem{johnson_using_1999}
Johnson, A.E., Hebert, M.: Using spin images for efficient object recognition
  in cluttered {3D} scenes. IEEE Transactions on Pattern Analysis and Machine
  Intelligence  \textbf{21}(5),  433--449 (May 1999). \doi{10.1109/34.765655}

\bibitem{liang_nonrigid_2018}
Liang, L., Wei, M., Szymczak, A., Petrella, A., Xie, H., Qin, J., Wang, J.,
  Wang, F.L.: Nonrigid iterative closest points for registration of {3D}
  biomedical surfaces. Optics and Lasers in Engineering  \textbf{100},
  141--154 (Jan 2018). \doi{10.1016/j.optlaseng.2017.08.005}

\bibitem{luthi_gaussian_2017}
L{\"u}thi, M., Gerig, T., Jud, C., Vetter, T.: Gaussian process morphable
  models. IEEE transactions on pattern analysis and machine intelligence
  \textbf{40}(8),  1860--1873 (2017)

\bibitem{von_luxburg_tutorial_2007}
von Luxburg, U.: A {Tutorial} on {Spectral} {Clustering}. arXiv:0711.0189 [cs]
  (Nov 2007)

\bibitem{ma_robust_2015}
Ma, J., Qiu, W., Zhao, J., Ma, Y., Yuille, A.L., Tu, Z.: Robust \${L}\_2e\$
  {Estimation} of {Transformation} for {Non}-{Rigid} {Registration}. IEEE
  Transactions on Signal Processing  \textbf{63}(5),  1115--1129 (Mar 2015).
  \doi{10.1109/TSP.2014.2388434}

\bibitem{ma_nonrigid_2019}
Ma, J., Wu, J., Zhao, J., Jiang, J., Zhou, H., Sheng, Q.Z.: Nonrigid {Point}
  {Set} {Registration} {With} {Robust} {Transformation} {Learning} {Under}
  {Manifold} {Regularization}. IEEE Transactions on Neural Networks and
  Learning Systems  \textbf{30}(12),  3584--3597 (Dec 2019).
  \doi{10.1109/TNNLS.2018.2872528}

\bibitem{ma_non-rigid_2016}
Ma, J., Zhao, J., Yuille, A.L.: Non-{Rigid} {Point} {Set} {Registration} by
  {Preserving} {Global} and {Local} {Structures}. IEEE Transactions on Image
  Processing  \textbf{25}(1),  53--64 (Jan 2016).
  \doi{10.1109/TIP.2015.2467217}

\bibitem{madsen_closest_2020}
Madsen, D., Morel-Forster, A., Kahr, P., Rahbani, D., Vetter, T., L{\"u}thi,
  M.: A closest point proposal for {MCMC}-based probabilistic surface
  registration. In: European {Conference} on {Computer} {Vision}. pp. 281--296.
  Springer (2020)

\bibitem{maiseli_recent_2017}
Maiseli, B., Gu, Y., Gao, H.: Recent developments and trends in point set
  registration methods. Journal of Visual Communication and Image
  Representation  \textbf{46},  95--106 (Jul 2017).
  \doi{10.1016/j.jvcir.2017.03.012}

\bibitem{mcneill_part-based_2006}
McNeill, G., Vijayakumar, S.: Part-{Based} {Probabilistic} {Point} {Matching}.
  In: 18th {International} {Conference} on {Pattern} {Recognition} ({ICPR}'06).
  vol.~2, pp. 382--386 (Aug 2006). \doi{10.1109/ICPR.2006.916}

\bibitem{mcneill_probabilistic_2006}
McNeill, G., Vijayakumar, S.: A {Probabilistic} {Approach} to {Robust} {Shape}
  {Matching}. In: 2006 {International} {Conference} on {Image} {Processing}.
  pp. 937--940 (Oct 2006). \doi{10.1109/ICIP.2006.312629}

\bibitem{morel-forster_probabilistic_2018}
Morel-Forster, A., Gerig, T., L{\"u}thi, M., Vetter, T.: Probabilistic fitting
  of active shape models. In: International {Workshop} on {Shape} in {Medical}
  {Imaging}. pp. 137--146. Springer (2018)

\bibitem{myronenko_point_2010}
Myronenko, A., Song, X.: Point {Set} {Registration}: {Coherent} {Point}
  {Drift}. IEEE Transactions on Pattern Analysis and Machine Intelligence
  \textbf{32}(12),  2262--2275 (Dec 2010). \doi{10.1109/TPAMI.2010.46}

\bibitem{myronenko_non-rigid_2007}
Myronenko, A., Song, X., Carreira-Perpi{\~n}{\'a}n, M.{\'A}.: Non-rigid point
  set registration: {Coherent} {Point} {Drift}. In: Sch{\"o}lkopf, B., Platt,
  J.C., Hoffman, T. (eds.) Advances in {Neural} {Information} {Processing}
  {Systems} 19, pp. 1009--1016. MIT Press (2007)

\bibitem{rasmussen_gaussian_2006}
Rasmussen, C.E., Williams, C.K.I.: Gaussian processes for machine learning.
  Adaptive computation and machine learning, MIT Press, Cambridge, Mass (2006)

\bibitem{robert_monte_2013}
Robert, C., Casella, G.: Monte {Carlo} statistical methods. Springer Science \&
  Business Media (2013)

\bibitem{rusu_fast_2009}
Rusu, R.B., Blodow, N., Beetz, M.: Fast {Point} {Feature} {Histograms} ({FPFH})
  for {3D} registration. In: 2009 {IEEE} {International} {Conference} on
  {Robotics} and {Automation}. pp. 3212--3217 (May 2009).
  \doi{10.1109/ROBOT.2009.5152473}

\bibitem{schonborn_markov_2017}
Sch{\"o}nborn, S., Egger, B., Morel-Forster, A., Vetter, T.: Markov {Chain}
  {Monte} {Carlo} for {Automated} {Face} {Image} {Analysis}. Int J Comput Vis
  \textbf{123}(2),  160--183 (Jun 2017)

\bibitem{steinke_kernels_2008}
Steinke, F., Sch{\"o}lkopf, B.: Kernels, regularization and differential
  equations. Pattern Recognition  \textbf{41}(11),  3271--3286 (Nov 2008).
  \doi{10.1016/j.patcog.2008.06.011}

\bibitem{umeyama_least-squares_1991}
Umeyama, S.: Least-{Squares} {Estimation} of {Transformation} {Parameters}
  {Between} {Two} {Point} {Patterns}. IEEE Transactions on Pattern Analysis and
  Machine Intelligence  \textbf{13}(04),  376--380 (Apr 1991).
  \doi{10.1109/34.88573}

\bibitem{wang_robust_2017}
Wang, D., Xue, J., Cui, D., Zhong, Y.: A robust submap-based road shape
  estimation via iterative {Gaussian} process regression. In: 2017 {IEEE}
  {Intelligent} {Vehicles} {Symposium} ({IV}). pp. 1776--1781 (Jun 2017).
  \doi{10.1109/IVS.2017.7995964}

\bibitem{woodbury_inverting_1950}
Woodbury, M.A.: Inverting modified matrices. Statistical Research Group (1950)

\bibitem{zhu_review_2019}
Zhu, H., Guo, B., Zou, K., Li, Y., Yuen, K.V., Mihaylova, L., Leung, H.: A
  review of point set registration: {From} pairwise registration to groupwise
  registration. Sensors  \textbf{19}(5), ~1191 (2019)

\bibitem{zou_non-rigid_2007}
Zou, G., Hua, J., Muzik, O.: Non-rigid surface registration using spherical
  thin-plate splines. In: International {Conference} on {Medical} {Image}
  {Computing} and {Computer}-{Assisted} {Intervention}. pp. 367--374. Springer
  (2007)

\end{thebibliography}
}

\clearpage 
\setcounter{page}{1}
{\Large \bf GiNGR Supplementary Material}
\appendix

\section{ICP-T Full Derivation}\label{app:icp-t}
The ICP-T minimization problem
\begin{equation}
    \argmin_{\tilde{U}} = \norm{
        \mat{ \lambda_s B \\ WI} \tilde{U} - \mat{ 0 \\ W\hat{U} }
    }_F^2
    \label{eq:deformationsquadraticWithW}
\end{equation}
To simplify the derivation, at first we set $W=I$ and ${\lambda_s=1}$, so the minimization instead becomes
\begin{equation}
    \argmin_{\tilde{U}} = \norm{
        \mat{ B \\ I} \tilde{U} - \mat{ 0 \\ \hat{U} }
    }_F^2
    \label{eq:deformationsquadraticWithoutW}
\end{equation}
Then we expand it into the least square solution:
\begin{equation}
\tilde{U}=(B^{T}B+I)^{-1}\hat{U}  
\end{equation}
Here we recognize that $L=B^{T}B$ is the Laplacian matrix and ${\hat{U}=X_c-X_R}$. We now make use of the Woodbury matrix identity
\begin{equation}\label{eq:woodburry}
(\mathcal{A}+\mathcal{U}\mathcal{C}\mathcal{V}^{T})^{-1} =
\mathcal{A}^{-1}-\mathcal{A}^{-1}\mathcal{U}(\mathcal{C}^{-1}+\mathcal{V}^{T}\mathcal{A}^{-1}\mathcal{U})^{-1}\mathcal{V}^{T}\mathcal{A}^{-1}.
\end{equation}
We set $\mathcal{A}^{-1}=L^{\dagger}=K$, and $\mathcal{U}$, $\mathcal{C}$, and $\mathcal{V}$ to identity matrices. Note, the Laplacian matrix $L$ does not have full rank and therefore we use the pseudo inverse. We can rewrite 
\begin{align*}
    \tilde{U} & = (L+ I)^{-1}\hat{U}\\
    \tilde{U} & = (K-K(K+ I)^{-1}K)\hat{U}
\end{align*}
The above equation is then set equal to the mean of GPR \cref{eq:gp-reg-mean-simple} when choosing the inverse Laplacian as the kernel. Then we can show that under these assumptions $u=\hat{U}$ holds, and hence both methods lead to the same prediction.
\begin{equation}
    \label{eq:icptProofBase}
    K(K+I)^{-1}u = (K-K(K+I)^{-1}K)\hat{U}
\end{equation}
Multiply with $(K+ I)K^{-1}$ on both sides
\begin{align}\label{eq:icptProofBaseNextStep}
\begin{split}
    u &= (K+I)K^{-1}(K-K(K+I)^{-1}K)\hat{U}\\
    u & = ((K+I)-K)\hat{U}\\
    u & = \hat{U}
\end{split}
\end{align}
which means that
\begin{equation}
    \tilde{U}=K(K+I)^{-1}\hat{U}.
\end{equation}

\subsection{Inclusion of Weighting Matrix W, $\lambda_s^2$ and $\sigma^2$}\label{appsub:weight-matrix}
We now focus on what the $W$ matrix, stiffness $\lambda_s$ and the noise assumption $\sigma^2$ variables change in the derivation. So we start out with the minimization problem in \cref{eq:deformationsquadraticWithW}
\begin{align*}
    \tilde{U} & = (L+\lambda_s^2 IW^{T}IW)^{-1}W^{T}IW\hat{U}\\
    \tilde{U} & = (K-KW^{T}(\sigma^2I+WKW^T)^{-1}WK)W\hat{U}
\end{align*}
Continuing the derivation similar to \cref{eq:icptProofBase}, this time equating to the GPR: $KW^T(\sigma^2 + WKW^T)^{-1}Wu$ extended with $W$, leads to
\begin{align}
\begin{split}
    KW^T(\sigma^2I + WKW^T)^{-1}Wu = \\ \bigl( \frac{1}{\lambda_s^2} K - \frac{1}{\lambda_s^2}KW(I + \frac{1}{\lambda_s^2}WKW)^{-1}\frac{1}{\lambda_s^2}WKW \bigr) W\hat{U}
\end{split}
\end{align}
Using $(I + \frac{1}{\lambda_s^2}WKW)^{-1} = \lambda_s^2 (\lambda_s^2I + WKW)^{-1}$ and also $W=WW=W^T$, as well as assuming $\lambda_s = \sigma$, the same approach as in \cref{eq:icptProofBaseNextStep} results in
\begin{align}
\begin{split}
    Wu &= (\sigma^2I + WKW)WK \bigl( \frac{1}{\lambda_s^2} K - \\
    &\, \qquad \frac{1}{\lambda_s^2}KW(\lambda_s^2I + WKW)^{-1}WKW \bigr)W\hat{U} \\
    Wu &= \bigl( \frac{1}{\lambda_s^2} (\sigma^2I + WKW)W - \frac{1}{\lambda_s^2} WKW \bigr)W\hat{U} \\
    Wu &= W\hat{U}
\end{split}
\end{align}
Therefore, the stiffness term $\lambda_s$ and the uncertainty assumption $\sigma$ are describing the same even when including $W$.
When comparing this to \cref{eq:gp-reg-mean-full}, we see that $WKW^T$ is the kernel function, but with all the unknown items removed. The same goes for $\hat{U}$ which now only contains the points that we have observed values for, the rest will be $0$. Likewise, $KW$ is the kernel matrix between all pairs of predicted points and the observed points, where the correlation between predicted points and unobserved points has been set to $0$. 

\section{ICP-A Full Derivation}\label{app:icp-a}
The ICP-A minimization problem
\begin{equation}
    \argmin_{\mathcal{M}} = \norm{
        \mat{ \lambda_s B \otimes G \\ WD} \mathcal{M} - \mat{ 0 \\ WX_c }
    }_F^2
\end{equation}
with $G$ being the $\diag([ 1,1,1,\gamma ]^T)$ and $\gamma$ depending on the units of the data. Again, we begin with the ICP-A energy optimization terms expanded to the least square solution. For simplicity we have set $W=I$ and $\lambda_s=1$. The inclusion of $W$ and $\lambda_s$ leads to the same outcome as shown in \cref{appsub:weight-matrix}.
\begin{align}
\begin{split}
\mathcal{M}=&( B^{T}B \otimes G^TG + D^TD)^{-1}D^TX_c \\
\mathcal{M}=&( L + D^TD)^{-1}D^TX_c 
\end{split}
\end{align}
We now make use of the Woodbury matrix identity \cref{eq:woodburry} and set $\mathcal{U}=D^T$, $\mathcal{V}=D$, $\mathcal{C}=I$ and $\mathcal{A}^{-1}=L^{\dagger}=K$.
\begin{equation}
    \mathcal{M}=(KD^T-KD^T ( I + DKD^T)^{-1}DKD^T)X_c
\end{equation}
As $\mathcal{M}$ is the stacked affine transformation matrices per point, we then multiply the solution with $D$ to get the updated point locations:
\begin{equation}
    \hat{X}_R=(DKD^T-DKD^T ( I + DKD^T)^{-1}DKD^T)X_c
\end{equation}
Setting $\mathcal{K}=DKD^T$ leads to
\begin{equation}
    \hat{X}_R=(\mathcal{K}-\mathcal{K} ( I + \mathcal{K})^{-1}\mathcal{K})X_c
\end{equation}
Similar to in the ICP-T derivation, we set the predicted point locations $\hat{X}_R$ equal to the GPR mean \cref{eq:gp-reg-mean-simple} using $\mathcal{K}$ as the kernel. Then we solve for the observation field $\hat{U}$.
Notice, in the ICP-A formulation, the updated locations of the points are predicted directly and not a deformation field $\tilde{U}$ that warps the reference. We can get the updated point locations in the standard GPMM formulation by adding the predicted mean deformation field to the reference point locations $X_R$,

\begin{equation}
    (\mathcal{K}-\mathcal{K} ( I + \mathcal{K})^{-1}\mathcal{K})X_c=X_R+\mathcal{K}(I + \mathcal{K})^{-1}\hat{U}.
\end{equation}

We now isolate $\hat{U}$ by multiplying both sides with ${(\mathcal{K}+I)\mathcal{K}^{-1}}$
\begin{multline}
    (((\mathcal{K}+I)\mathcal{K}^{-1})(\mathcal{K}-\mathcal{K} ( I + \mathcal{K})^{-1}\mathcal{K}))X_c = \\(\mathcal{K}+I)\mathcal{K}^{-1}X_R+\hat{U}
\end{multline}
\begin{multline}
        ((\mathcal{K}+I)-\mathcal{K})X_c = (X_R+ \mathcal{K}^{-1}X_R)+\hat{U}
\end{multline}
\begin{equation}
        X_c  - (X_R+ \mathcal{K}^{-1}X_R) =\hat{U}
\end{equation}
The additional term $\mathcal{K}^{-1}X_R$ (in comparison to ICP-T) is to adjust for the ICP-A algorithm defining its closest point deformations as affine transformations.
\end{document}